\pdfoutput=1

\documentclass[11pt]{article}

\usepackage[final]{acl}

\usepackage{times}
\usepackage{latexsym}
\usepackage{tikz}
\usepackage{subcaption}
\usepackage{hhline}
\usepackage{makecell}

\usepackage[T1]{fontenc}

\usepackage[utf8]{inputenc}
\usepackage{amsmath}
\usepackage{xspace}
\usepackage{colortbl}
\usepackage{rotating}

\usepackage{microtype}

\usepackage{inconsolata}
\usepackage{booktabs} 

\definecolor{skyblue}{RGB}{135,206,235}

\usepackage{graphicx}

\newcommand{\SMART}{SMART filtering\xspace}

\newcommand\blfootnote[1]{%
  \begingroup
  \renewcommand\thefootnote{}\footnote{#1}%
  \addtocounter{footnote}{-1}%
  \endgroup
}

\title{Improving Model Evaluation using SMART Filtering of Benchmark Datasets}

\author{%
Vipul Gupta\textsuperscript{$1,2,\dagger$} \enspace Candace Ross\textsuperscript{$1$} \enspace David Pantoja\textsuperscript{$3$} \enspace Rebecca J. Passonneau\textsuperscript{$2$} \\ \enspace \textbf{Megan Ung}\textsuperscript{$1$} \enspace \textbf{Adina Williams}\textsuperscript{$1$}
\\
{\textsuperscript{$1$} {FAIR, Meta AI}}  \quad
{\textsuperscript{$2$} {Pennsylvania State University} }\quad  
{\textsuperscript{$3$} {University of California, Berkeley
} }\quad  \\
\quad\\
\normalsize{\tt vkg5164@psu.edu, } \normalsize{\tt \{ccross, meganu, adinawilliams\}@meta.com}\\
}

\begin{document}
\maketitle
\begin{abstract}

One of the most challenging problems facing NLP today is evaluation. Some of the most pressing issues pertain to benchmark saturation, data contamination, and diversity in the quality of test examples. To address these concerns, we propose \textit{Selection Methodology for Accurate, Reduced, and Targeted (SMART) filtering}, a novel approach to select a high-quality subset of examples from existing benchmark datasets by systematically removing less informative and less challenging examples. Our approach applies three filtering criteria, removing (i) easy examples, (ii) data-contaminated examples, and (iii) examples that are similar to each other based on distance in an embedding space. We demonstrate the effectiveness of \SMART on three multiple choice QA datasets, where our methodology increases efficiency by reducing dataset size by 48\% on average, while increasing Pearson correlation with rankings from ChatBot Arena, a more open-ended human evaluation setting. Our method enables us to be more efficient, whether using \SMART to make new benchmarks more challenging or to revitalize older datasets, while still preserving the relative model rankings.

\end{abstract}

\blfootnote{$\dagger$ This work was done during an internship at FAIR, Meta.}

\section{Introduction}


\begin{figure}[t]
\begin{center}
\includegraphics[width=0.97\linewidth]{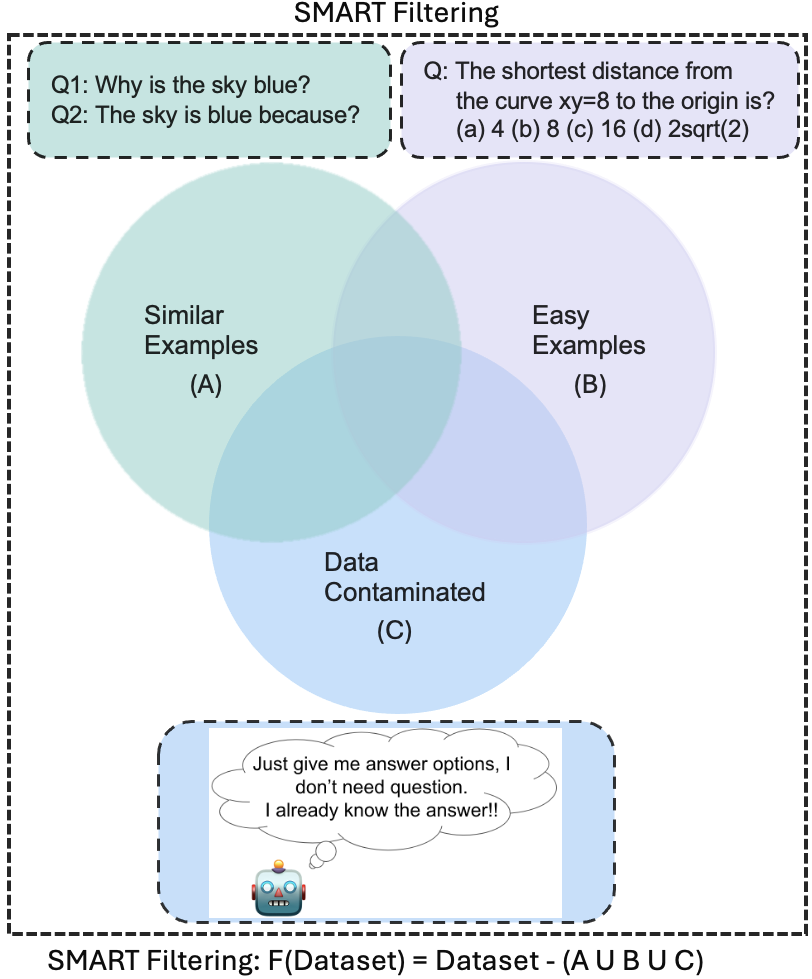}
\end{center}
\vspace{-1em}
  \caption{
  This figure illustrates our methodology. We remove easy, data contaminated, and/or similar examples from datasets to find a high-quality subset. The examples selected are from MMLU dataset.}
\vspace{-1em}
\label{fig:main}
\end{figure}

\begin{table*}[ht]
\centering
\small
\begin{tabular}{lrrrrrrr}
\specialrule{1.5pt}{1pt}{0.3pt} 
\multicolumn{2}{c}{\normalsize{\cellcolor[HTML]{8BC34A} Dataset}} & \multicolumn{4}{c}{\normalsize{\cellcolor[HTML]{87CEEB} Filtering Steps (in \%)}} & \multicolumn{2}{c}{\normalsize{\cellcolor[HTML]{C9C9A5} SMART Filtering}} \\ \midrule
Name & \# of examples & Easy & Data Contaminated & Similar & Prefiltering & \% Filtered & \# examples \\
\midrule
ARC & 3530 & 64.41 & 3.45 & 12.57 & 0.22 & \textbf{68.92} & \textbf{1097} \\
MMLU & 14042 & 35.01 & 4.37 & 3.31 & 7.61 & \textbf{43.02} & \textbf{8000} \\
CommonsenseQA & 1221 & 27.60 & 0.00 & 8.90 & 0.16 & \textbf{34.30} & \textbf{802}\\
\bottomrule
\end{tabular}
\caption{SMART filtering of MMLU, CommonsenseQA, and ARC results in leaner, more informative evaluations. }
\label{tab:dataset_stats}
\end{table*}

With the recent rise of interest in natural language generation and language modeling as a general purpose pretraining task, model benchmarking---evaluating several models on a standard test set to derive a strict ranking---has become much less straightforward. This is in part due to the increasing speed of evaluation dataset saturation \citep{kiela-etal-2021-dynabench, vania-etal-2021-comparing, ott-etal-2022-mapping}, whereby current state-of-the-art models are able to match human-level performance within the margin of error \cite{laskar-etal-2023-systematic, keeganblog}. 
This signals the need for new, potentially more challenging benchmarks; 
however, generating high quality human-annotated datasets is difficult \cite{rein2023gpqa}. It is time-consuming to devise new evaluation datasets, which can require training and coordinating human annotators, cleaning data, and assessing test validity \cite{sun2021analyzing, herde2021survey}. Moreover, the increasing speed of benchmark saturation leads to a difficult value proposition for benchmark creators who have only so much time to make important research contributions. 

As if the speed of progress in NLP benchmark saturation weren't enough of a challenge, the increasingly wide availability of text-generation systems has made the classical approach to collecting high quality, human-baselined evaluation data---crowd-sourcing---much less tenable. Since detecting model-generated text is an open area of research \citep{jawahar-etal-2020-automatic, dugan-etal-2023-real}, benchmark developers can no longer trust that a new, purportedly human-created evaluation dataset was indeed created by humans, and not humans using automatic tools as assistants. This state of affairs raises an important question about past evaluation datasets: if they weren't saturated, could they still help us rank models based on task performance? Can we recapture some of the utility in saturated evaluation datasets that appear to be saturated?

One common approach to dataset saturation is filtering \cite{pmlr-v119-bras20a}. Just as renovating an older home is generally more efficient than building a new home from scratch, dataset filtering can rejuvenate a high-quality older benchmark, without the need to rely on novel rounds of costly human data collection. Along these lines, we propose Selection Methodology for Accurate, Reduced and Targeted (SMART) filtering, a novel approach for finding a high-quality subset of evaluation benchmarks that doesn't need human intervention, illustrated in \autoref{fig:main}. 
We achieve this by algorithmically removing the low-quality examples from existing evaluation datasets, preserving the high-quality examples that are most informative and reliable as our basis for benchmarking. Our methodology applies three filtering criteria: 1) removing easy examples, 2) removing data-contaminated examples that are highly likely to have been leaked into the training datasets, and 3) removing similar examples.



All \SMART steps have their conceptual origin in dataset filtering, which can reduce the data size and hence compute costs \citep{treviso-etal-2023-efficient, mishra-sachdeva-2020-need}. In particular, we utilize filtering methods used for training-time deduplication \citep{lee-etal-2022-deduplicating, abbas2023semdedup, swayamdipta-etal-2020-dataset}, a technique often used during pretraining to increase efficiency. We contend that pretraining deduplication approaches will also be effective for benchmarking filtering. They can increase headroom on standard, already vetted, human-written datasets that have been saturated, and re-enable meaningful model rankings. 

Dataset filtering, if correctly applied (c.f., \citealt{phang-etal-2022-adversarially}), can retain or even improve benchmark dataset quality \citep{treviso-etal-2023-efficient}, meaning that \SMART might be able to contribute to addressing reliability issues arising from current day benchmarking practices \cite{gupta2024changing, alzahrani-etal-2024-benchmarks, sinha-etal-2021-unnatural}. Each of the steps in \SMART aims to improve usefulness and reliability by removing examples that are not informative for differentiating  models. 

\SMART could also be applied to new datasets at the earlier stages of the benchmark lifecycle before they become standards. When used before new benchmarks premiere, \SMART can ensure that the final version is as challenging, high quality, and efficient as possible.

We test \SMART on the classical multiple choice QA setting. We characterize the impact of \SMART using three datasets---ARC \cite{clark2018think}, MMLU \cite{hendrycks2021measuring} and CommonsenseQA \cite{talmor-etal-2019-commonsenseqa}. In Tables \ref{tab:dataset_stats} and \ref{tab:smart_ranking}, we show that our methodology increases efficiency without hits to leaderboard utility. For example, on the ARC dataset, we shrink the dataset by 68.9\%, increasing computational efficiency drastically, while observing practically no effect on the relative ranking of models. We also compare model rankings of a dataset with open-ended human evaluated ChatBot Arena \cite{chiang2024chatbot} rankings, and found that datasets using \SMART have a stronger correlation with ChatBot Arena indicating better correlation with human preferences.

\begin{table*}[htbp]
  \centering
  \scriptsize
  \setlength{\tabcolsep}{4pt}
 \begin{tabular}{l|cccccccccccc}
\toprule
    & \rotatebox{75}{Qwen2-72B-it} & \rotatebox{75}{Llama-3.1-70B-it} & \rotatebox{75}{Llama-3-70B-it} & \rotatebox{75}{Gemma-2-27b-it} & \rotatebox{75}{Phi-3.5-MoE-it} & \rotatebox{75}{Mixtral-8x22B-it} & \rotatebox{75}{Dbrx-it} & \rotatebox{75}{Yi-34B} & \rotatebox{75}{Llama-3-8B-It} & \rotatebox{75}{Qwen-7B-Chat} & \rotatebox{75}{Gemma-7b-it} \\
    \midrule
    ARC & 93.9 (1) & \cellcolor{skyblue} 93.4 (2) & \cellcolor{skyblue} 93.3 (3) & 92.5 (4) & 92.0 (5) & 91.6 (6) & 90.5 (7) & 90.2 (8) & 88.3 (9) & \cellcolor{skyblue} 76.4 (10) & \cellcolor{skyblue} 73.8 (11) \\
    ARC-SMART & 83.0 (1) & \cellcolor{skyblue} 81.9 (2) & \cellcolor{skyblue} 81.9 (2) & 78.8 (3) & 78.5 (4) & 76.2 (5) & 73.2 (6) & 72.4 (7) & 72.1 (8) & \cellcolor{skyblue} 51.8 (10) & \cellcolor{skyblue} 53.1 (9) \\\midrule
    MMLU & 84.1 (1) & 82.3 (2) & 80.3 (3) & 76.2 (6) & 78.7 (4) & 77.9 (5) & \cellcolor{skyblue} 73.2 (8) & \cellcolor{skyblue} 75.8 (7) & 66.4 (9) & 56.3 (10) & 51.7 (11) \\
    MMLU-SMART & 74.3 (1) & 71.4 (2) & 69.2 (3) & 63.9 (6) & 67.0 (4) & 65.3 (5) & \cellcolor{skyblue} 60.0 (7) & \cellcolor{skyblue} 62.4 (8) & 50.5 (9) & 41.5 (10) & 38.9 (11) \\\midrule
    CommonsenseQA & 88.5 (1) & 81.5 (3) & 83.3 (2) & \cellcolor{skyblue} 79.4 (6) & 80.9 (4) & 75.9 (9) & 78.6 (7) & \cellcolor{skyblue} 79.7 (5) & 76.5 (8) & 65.2 (11) & 68.8 (10) \\
    CommonsenseQA-SMART & 84.5 (1) & 74.1 (3) & 77.1 (2) & \cellcolor{skyblue} 71.9 (5) & 73.9 (4) & 67.2 (9) & 70.4 (7) & \cellcolor{skyblue} 71.8 (6) & 68.0 (8) & 55.7 (11) & 59.4 (10) \\
    \bottomrule
\end{tabular}
\caption{Model performance on ARC, MMLU, CommonsenseQA, and their SMART-filtered counterparts. Accuracy scores are accompanied by rankings in parentheses. Cells highlighted {\color{skyblue}\bf in blue} indicate models whose ranking shifted after applying SMART filtering. The notation `\textit{it}' refers to instruction-tuned versions of the model.}
\label{tab:smart_ranking}
\end{table*}

\section{Related Works}

Recently, many benchmark datasets  such as MMLU \cite{hendrycks2021measuring}, GSM8K \cite{cobbe2021training}, MATH \cite{hendrycks2021math} and GPQA \cite{rein2023gpqa} have been proposed to measure the capabilities of language models (LMs). However, recent works have highlighted issues that call into question the reliability of these datasets. Despite extensive efforts in dataset creation, annotation errors still persist \cite{gema-etal-2024-we, wang2024mmlu}. Recently, some works have shown that the accuracy of models on multiple-choice question datasets can change significantly by simply altering the order of answer options \cite{pezeshkpour-hruschka-2024-large, gupta2024changing, sugawara2020assessing, zong2024fool}. One proposal  suggests modifying the self-attention matrix to prevent answer options from paying attention to each other by masking their scores to zero \citep{mcilroyyoung2024setbasedpromptingprovablysolving}, but nonetheless, benchmark validity and gameability remains an open area of ongoing research.

Further complicating evaluation, some models have a prior bias towards a particular option id (e.g.\ `A') \cite{zheng2024large, wei-etal-2024-unveiling, zheng2023judging, li2024anchored, reif-schwartz-2024-beyond, ross-etal-2024-what}. Additionally, models can perform well above random chance even when question text is removed and only answer options are given, potentially highlighting possible data contamination issues \cite{balepur-etal-2024-artifacts, shah-etal-2020-expect}. Recent works have shown that replacing the correct option with ``None of the above'' leads to a drastic decline in performance across all models \citep{wang2024beyond, xu2024llms}. 
This calls into question the reliability of datasets and leaderboard ranking as the accuracy of a model may not be representative of its capabilities \cite{rottger2024political, raj2023semantic}. Under the assumption that leaked, gameable, spuriously correlated, and/or biased examples are often easier for models to do well on, we aim to help address such issues by filtering ``easier'' examples from benchmarks to increase reliability. 

Other works have explored related approaches to improve evaluation \cite{vivek-etal-2024-anchor, pmlr-v119-bras20a, gupta2023calm}. 
\citet{varshney-etal-2022-ildae} tested models only on difficult examples of a dataset, showing that evaluating on just 5\% examples can achieve a very high correlation with the full dataset. However, their approach requires retraining the models on various subsets of training data and uses humans to verify the filtered dataset. In the current NLP paradigm, access to training data is often unavailable for most models \cite{longpre2024consent} and even when accessible, it is very expensive to retrain models on those datasets. Contrasting with \citet{varshney-etal-2022-ildae},  our approach emphasizes computational efficiency and identifies a high-quality subset  that correlates well with the original dataset, without requiring access to training data or expensive human verification. 










\section{SMART Filtering Methodology}
Our methodology has three main filtering steps, each of which is applied independently; the order of filtering steps does not impact the final subset. 

All three steps in \SMART are aimed at deduplicating and making NLP leaderboarding more efficient. We contend that examples in the dataset that all tested models get correct with very high confidence are not useful for establishing model ranking (although they might be useful for data exploration or other things). Therefore, including such ``easy'' examples merely increases the computation cost without providing ability to distinguish between models in leaderboard settings \cite{varshney-etal-2022-ildae}. Relatedly, examples that are present in pretraining datasets should not be used for model testing \citep{elangovan-etal-2021-memorization}. Leaked examples can give some models unwanted and difficult-to-interpret advantages when they are used for a leaderboard \citep{jiang2024investigating, ravaut2024much}. Finally, examples that are overly similar to one another effectively contribute less information about model performance than examples that meaningfully differ. Example overlap can also contribute towards idiosyncratically favoring particular kinds of information and potentially even double counting. In what follows, we will describe each step in the \SMART pipeline, to enable others to apply our methodology to benchmarks from any NLP task.

\subsection{Pre-filtering}
Before applying the \SMART steps, we apply two pre-filtering criteria. First, we eliminated exact match duplicate examples, keeping only one instance of each unique example. Copied examples in a dataset lead to overestimation of model performance \cite{matatov2022dataset}, and thus they should be removed. We leave this step as a ``pre-filtering'' step, as most evaluation datasets have already been filtered for exact-match copies by their creators, usually making this step unnecessary. We provide more details in Appendix \ref{subsec:exact_match}.

Our second prefiltering criteria is the removal of anomalous example subsets, which we define as examples that are significantly different in form from the rest of the dataset, as this makes them incommensurable and hard to interpret. These subsets are often low quality and decrease the reliability of the dataset. For instance, the moral scenarios subset in MMLU, a subset of the ETHICS dataset \citep{hendrycks-etal-2020-aligning}, is notably different from other categories. Moreover, automating machine morality has been argued to have deep conceptual issues \citep{talat-etal-2022-machine}, raising questions about whether such a task should be included in a general purpose benchmark. 
We removed such subsets.

\subsection{Filtering Easy Examples}

In this step, we focus on identifying and removing the easy examples from the dataset \cite{gururangan-etal-2018-annotation, rodriguez-etal-2021-evaluation}. We define easy examples based on the agreement between top-performing open-source models from Open LLM leaderboard \citep{open-llm-leaderboard}. Specifically, we consider an example to be easy if there is unanimous agreement among all top-performing models, with each model answering the example correctly with high confidence (greater than $0.8$). This approach finds examples that are consistently easy across different model architectures and models trained on different datasets. By removing these examples, we also reduce the computational cost of running inference on the resulting dataset.

The rationale behind removing these examples is that such agreement between various models indicates that they do not help in providing meaningful insights into relative model capabilities as all models answer them correctly. However, to ensure future models don't lose the ability to answer these easy examples, we elected to retain a random 10\% of the easy examples in the \SMART (and associated them with metadata to enable future analysis). This helps us to adhere more closely to the distribution of the original benchmarks, while still achieving significant computational efficiency.

\subsection{Filtering Data Contaminated Examples}
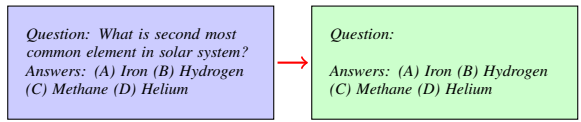
\begin{figure}[b]
\centering
\begin{tikzpicture}
  \draw[fill=blue!20] (0,0) rectangle (3.5,1.5);
  \node[align=left, text width=3cm, font=\tiny\itshape] at (1.75,0.75) {
    Question: What is second most common element in solar system? \\
    Answers: (A) Iron (B) Hydrogen (C) Methane (D) Helium
  };
  \draw[fill=green!20] (4,0) rectangle (7.5,1.5);
  \node[align=left, text width=3cm, font=\tiny\itshape] at (5.75,0.75) {
    Question: \textcolor{green!20}{What is second most common element in solar system?}\\
    Answers: (A) Iron (B) Hydrogen (C) Methane (D) Helium
  };
  \draw[->, thick, red] (3.55,0.75) -- (3.95,0.75);
\end{tikzpicture}
\vspace{-0.5em}
\caption{We remove question context and give answer-only prompt. If all models still predict the correct answer with high probability ($>0.8$), then we categorize that example as data contaminated.}
\label{fig:data_cont}
\end{figure}

In this step, we identify and remove examples that are likely to be data contaminated, i.e. present in training data of models \cite{deng2023benchmark, elazar2024whats, magar-schwartz-2022-data}. This is important because evaluating models on examples they have already seen during training can lead to inflated numbers, giving a false sense of higher model performance.
Detecting data contamination is inherently challenging as also highlighted in recent studies \cite{duan2024membership, singh2024evaluation}, and many promising approaches require access to training data \cite{jiang2024investigating}.
To identify data contaminated examples, we follow an approach inspired by \citet{balepur-etal-2024-artifacts} and \citet{balepur-rudinger-2024-large} as shown in Figure \ref{fig:data_cont}. In this approach, we do not need access to training data. Specifically, we modify the prompt to the model by removing the question text and presenting only the answer choices.
This approach challenges the model to select the correct answer in an artificial setting without the context of the question itself.

In this step, we again take agreement between top-performing models.
We employ a stringent criterion: an example is deemed contaminated only if \textit{all} top \textit{n} open-source models answer it correctly without any context and with high confidence (greater than $0.8$). This conservative approach prioritizes precision over recall to ensure that the examples removed are truly contaminated. We recognize that this strict criterion may underestimate the number of contaminated examples. As the research in detecting data contamination improves, we plan to incorporate more sophisticated techniques that can detect contamination more effectively while balancing precision and recall. 


\subsection{Filtering Similar Examples}\label{sec:threshold}
This filtering step identifies highly similar examples within the dataset to prevent redundant evaluations and avoid potential bias towards models that perform well on these particular closely related examples. 
We use an embedding-based approach, representing each example (question and answer choices) in high dimensional space. We experimented with two prominent embedding methods: SentenceBert \cite{reimers-gurevych-2019-sentence}, which encodes each input into a $384$-dimensional vector, and LLM2Vec \cite{behnamghader2024llm2vec}, which modifies decoder-only LMs for sentence representation.
Both approaches use bidirectional attention to capture rich contextualized representation. However, due to widespread usage in various applications \cite{su-etal-2023-one, khattab2020colbert}, and lower computational requirements, we prioritized SentenceBert in \SMART.

\begin{figure}[b]
\begin{center}
\includegraphics[width=\linewidth]{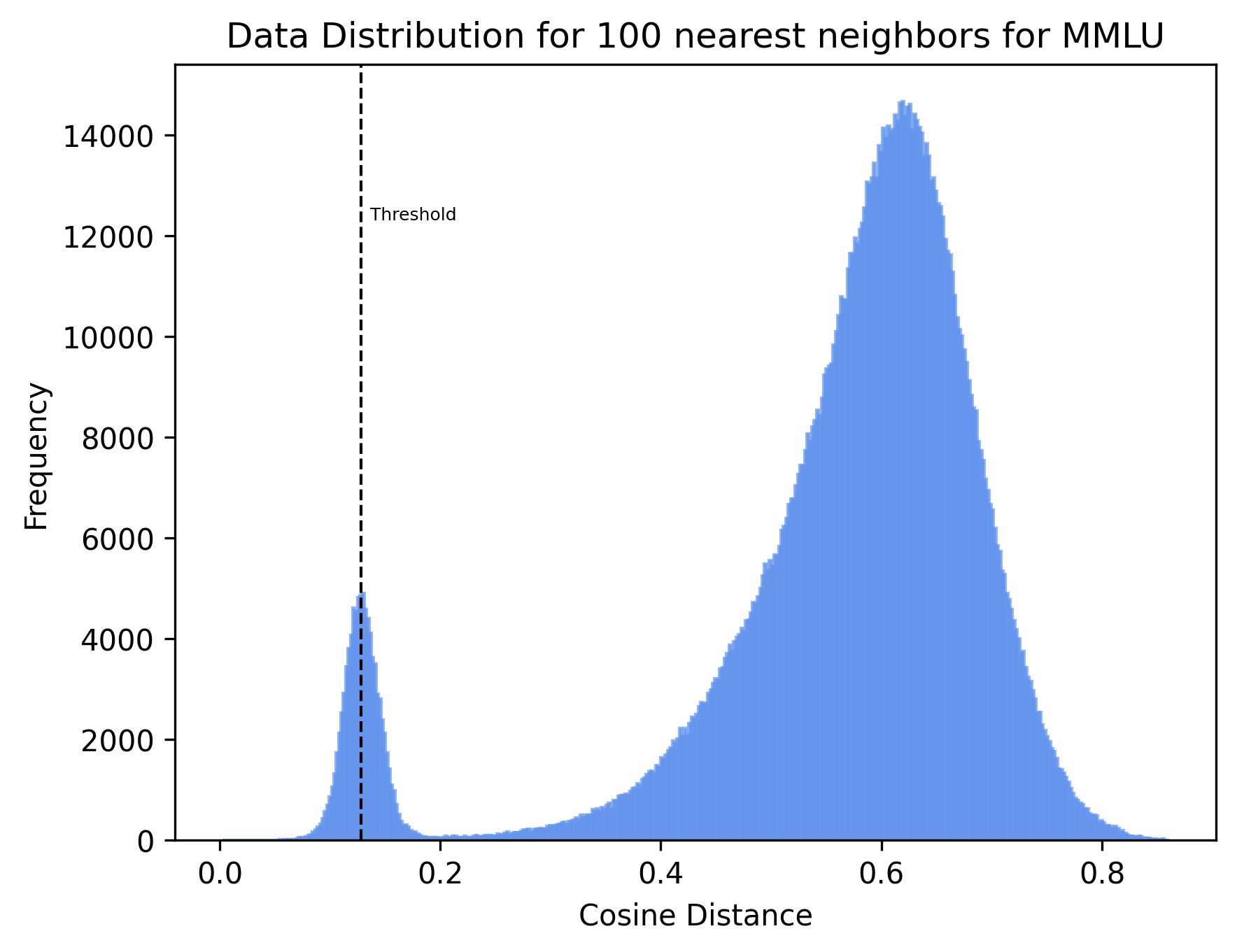}
\end{center}
\vspace{-1em}
\caption{
Cosine distances between SentenceBert embeddings for MMLU examples. The black vertical line is the threshold for identifying similar example pairs.
}
\label{fig:sbert_mmlu}
\end{figure}

We compute the cosine distance between each pair of example embeddings. 
Cosine distance is defined as the cosine similarity subtracted from $1$ and ranges from $0$ to $2$, with $0$ signifying maximum similarity between two embeddings. To determine an appropriate threshold, $\delta$, for identifying similar pairs, we analyze the distribution of cosine distances as shown in Figure \ref{fig:sbert_mmlu}. We employ kernel density estimation to locate the first local maximum of this distribution, capturing the point at which the similarity begins to decrease significantly. This data-driven approach to selecting $\delta$ allows us to adapt to the specific characteristics of each dataset. Pairs with distance lower than $\delta$ (indicated by the black vertical dotted line in Figure \ref{fig:sbert_mmlu}) are considered similar and grouped into clusters. 
To maintain dataset distribution, we randomly remove half of the examples from each identified cluster. 
We provide more details on threshold selection in Appendix \ref{subsec:threshold_analysis}.

\begin{figure*}[t]
  \centering
  \subfloat[ARC]{\includegraphics[width=0.32\textwidth]{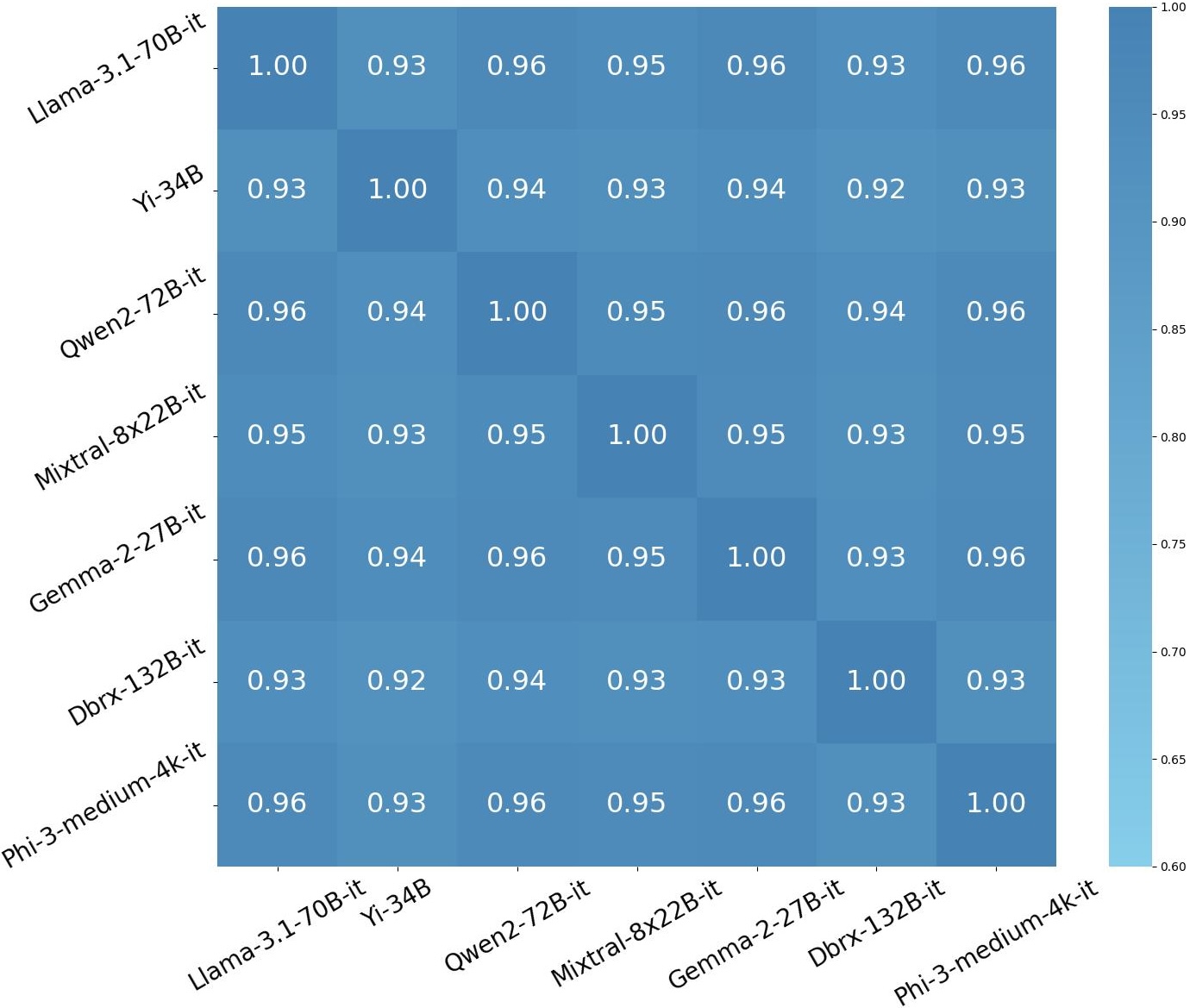}} \hfill
  \subfloat[MMLU]{\includegraphics[width=0.32\textwidth]{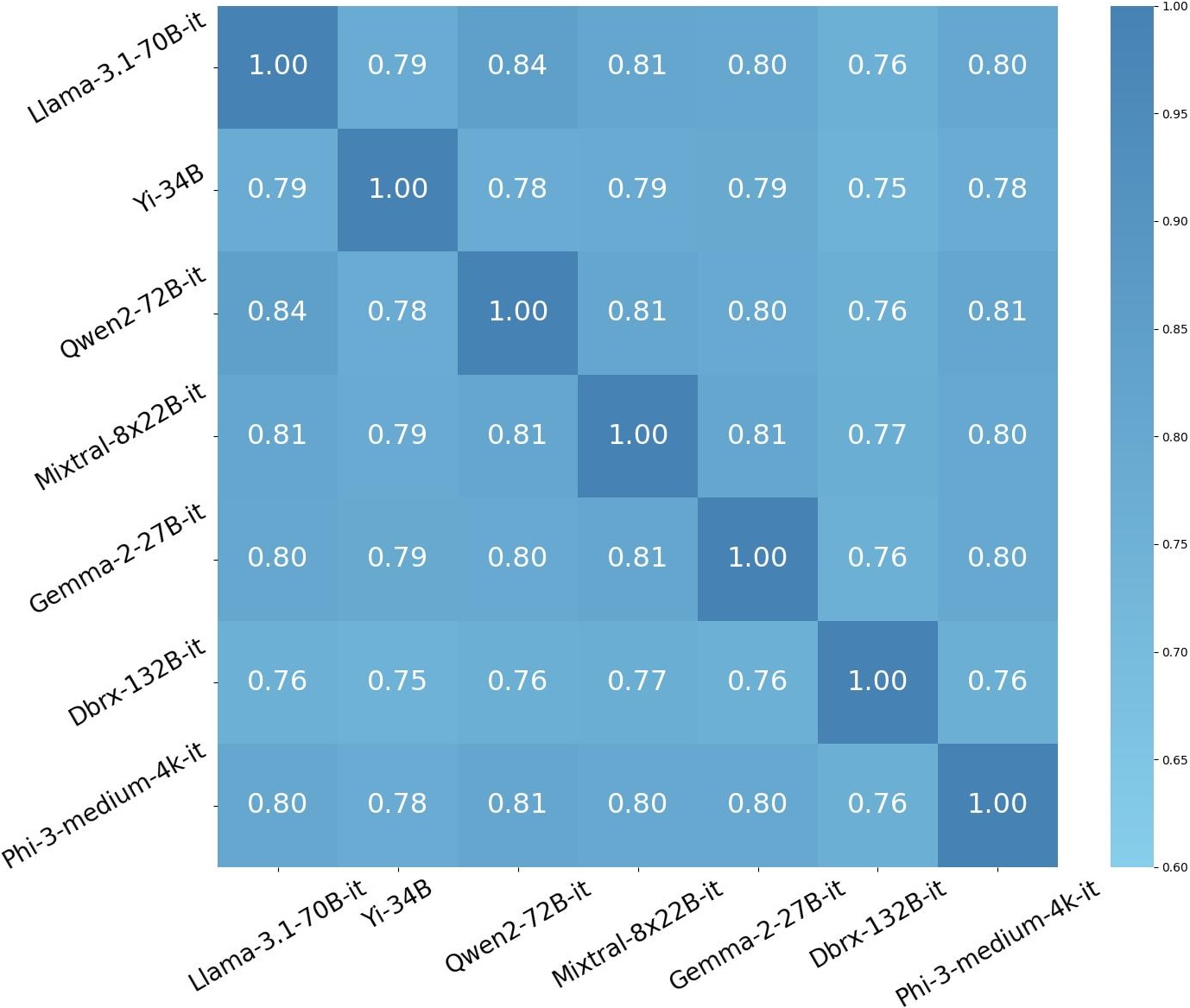}} \hfill
  \subfloat[Commonsense QA]{\includegraphics[width=0.32\textwidth]{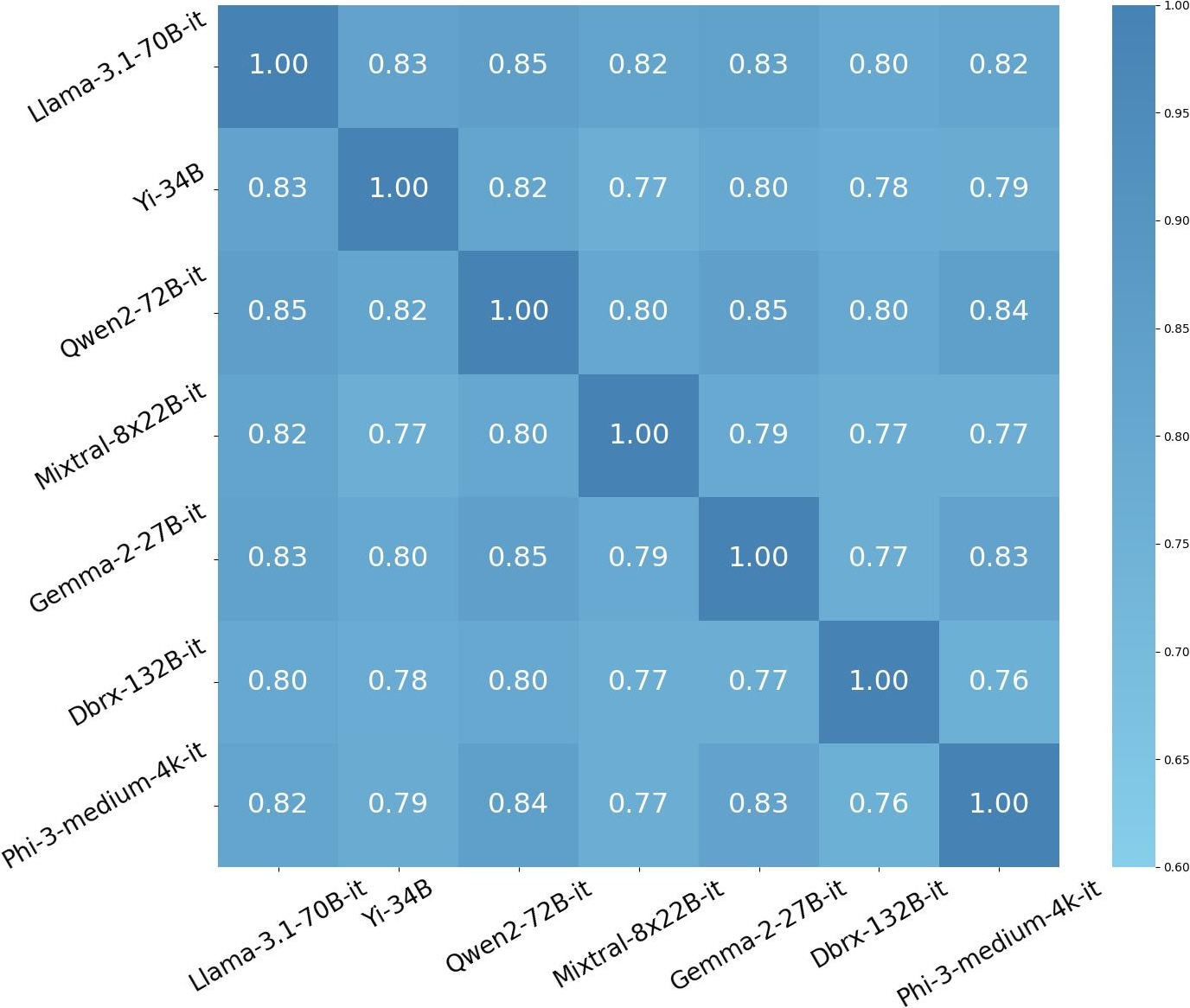}} \\
  \subfloat[ARC-\SMART]{\includegraphics[width=0.32\textwidth]{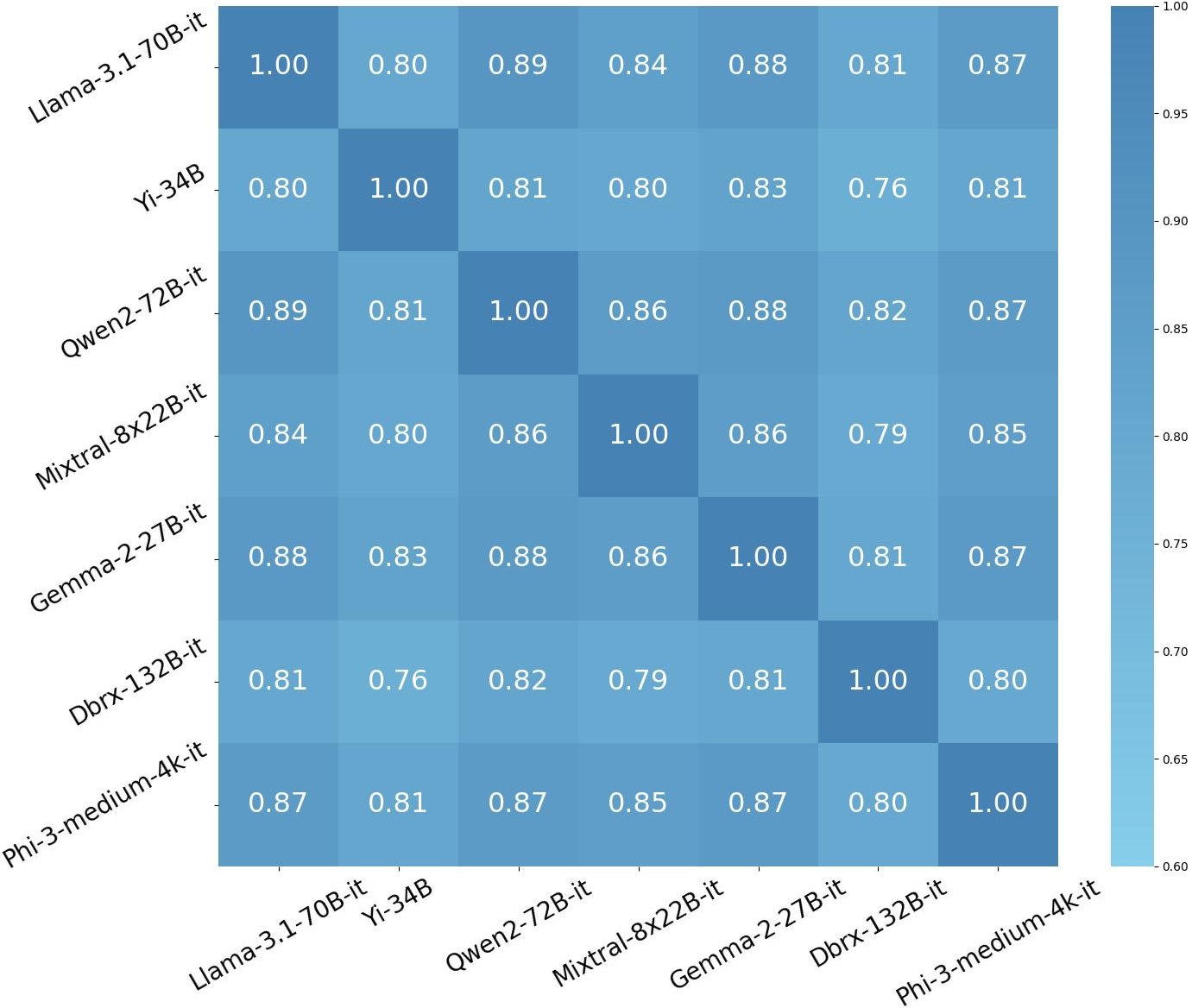}} \hfill
  \subfloat[MMLU-\SMART]{\includegraphics[width=0.32\textwidth]{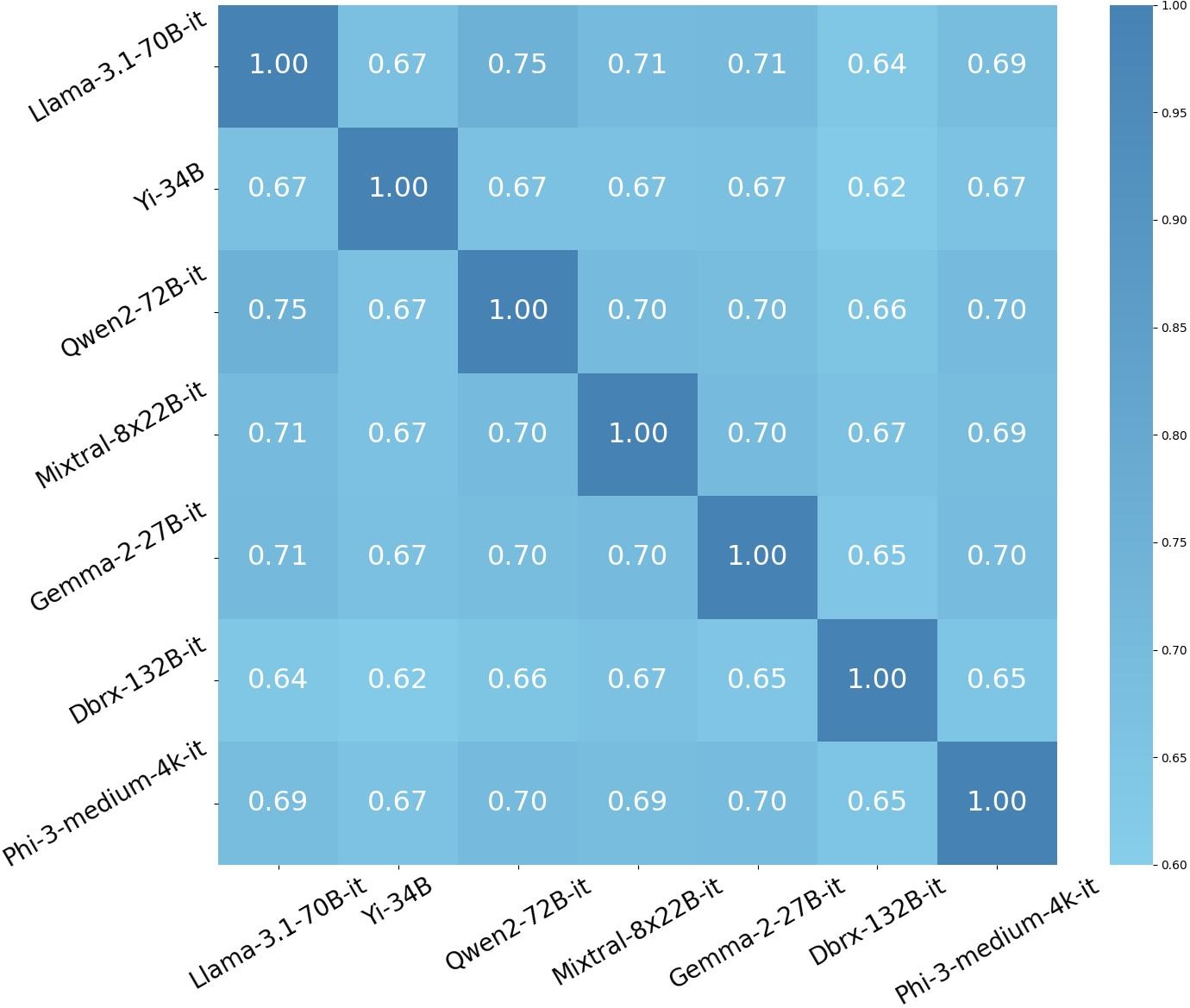}} \hfill
  \subfloat[Commonsense QA-\SMART]{\includegraphics[width=0.32\textwidth]{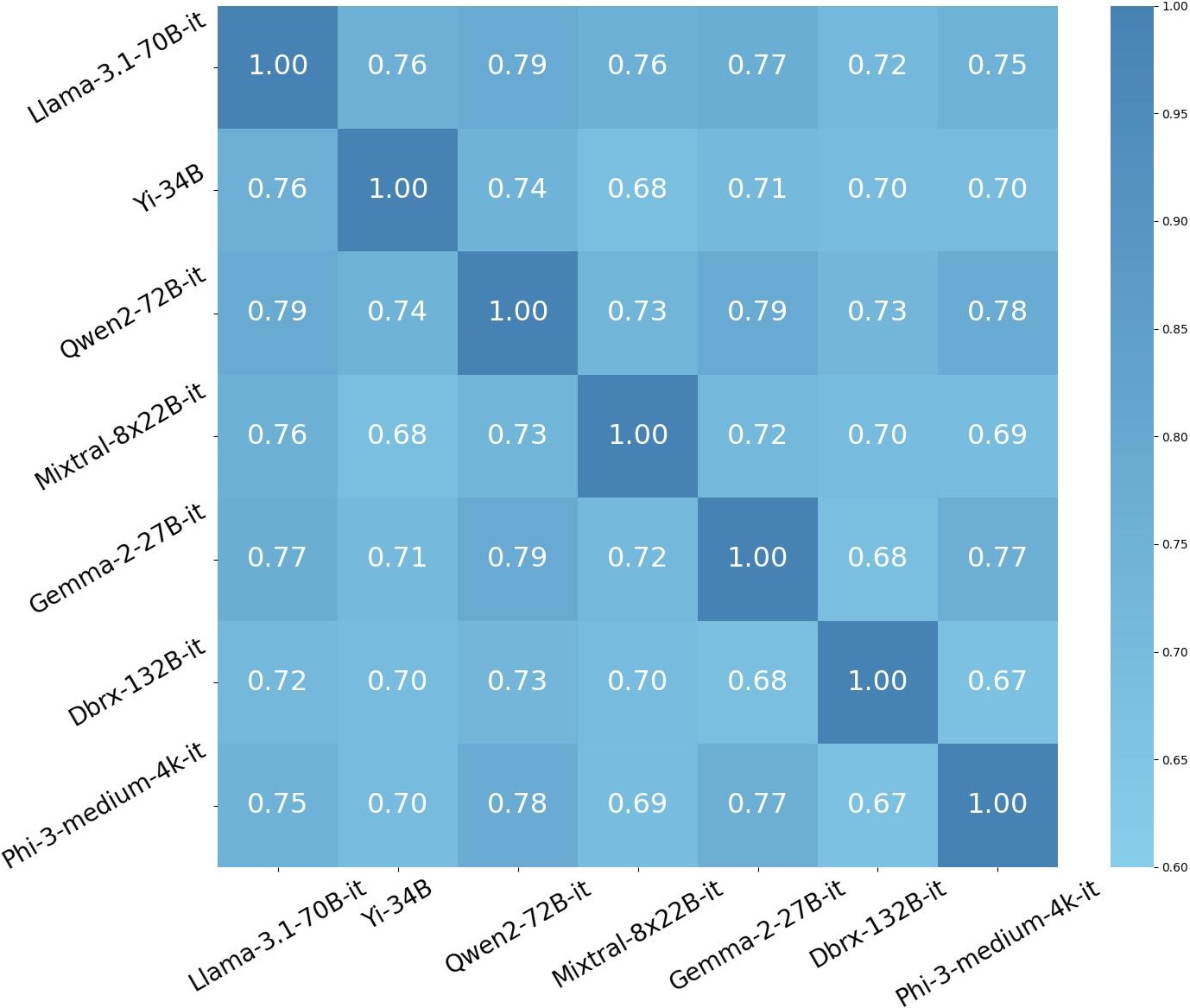}}
  \caption{Heatmap illustrating the degree of agreement between model predictions for ARC, MMLU and Commonsense QA datasets, as well as their \SMART variants. Notably, \SMART leads to a decrease in inter-model agreement, indicating that our approach helps in better differentiating the capabilities of the models.}
  \label{fig:heatmaps}
\end{figure*}

\section{Results: Efficiency \& Informativeness}

In this section, we present the results of applying \SMART on popular datasets, focusing on three well-established multiple-choice question answering datasets: ARC \cite{clark2018think}, MMLU \cite{hendrycks2021measuring} and CommonsenseQA \cite{talmor-etal-2019-commonsenseqa}. While we show results on MCQA datasets, our methodology is generic and can be applied to various dataset types. Given that all other classification tasks can be, and often are, recast into MCQA format, it strikes us as likely that SMART-filtering would work just as well for classical NLU tasks (like NLI, coreference resolution, sentiment analysis) without needing novel strategies. Although these tasks have conceptual differences, with respect to the format, these variations boil down to having different numbers of answer options. Going into specifics of why we selected these datasets, we selected MMLU and ARC datasets because of their widespread adoption in benchmarking by state-of-the-art models \cite{meta-llama3.1, google2023gemma} and their coverage of diverse topics and difficulties. These two datasets are 4-choice MCQA, so to broaden the scope, we included CommonsenseQA, which offers 5-choice MCQA. We think our inclusion of different numbers of MCQA answer options between 4 and 5 gives some indication that the methodology will be adaptable to smaller numbers of answer options (2 or 3) too.
We release our code and provide \SMART version of each dataset\footnote{ Link to our codebase: \url{https://github.com/facebookresearch/ResponsibleNLP/tree/main/SMART-Filtering}.}.

To identify easy and data contaminated examples, we evaluate 7 open source models from the top of the Open LLM leaderboard \citep{open-llm-leaderboard} on each QA dataset: Llama 3.1-70B \citep{meta-llama3.1}, Phi-3-14B \citep{abdin2024phi}, Yi-34B \citep{ai2024yi}, Qwen2-72B \citep{qwen2}, Mixtral-8x22B \citep{mistral2023mixtral}, Gemma2-27B \citep{google2023gemma}, and DBRX-132B \citep{databricks2023dbrx}. We use instruction tuned (`instruct') versions when available, as they should have better question-answering capabilities.

Moreover, as these models came from different organizations, they were likely trained on different datasets. Ideally, this minimizes the risk of bias towards any particular model when identifying easy and contaminated examples (see Section~\ref{sec:num_model} for more discussion on this). We used open-source models, as we aim to develop a methodology that is accessible to any AI practitioner with sufficient computation resources, thereby avoiding the substantial closed-source API costs.

The effectiveness of \SMART is shown in \autoref{tab:dataset_stats} as the percentage of examples filtered for ARC,\footnote{Because we excluded the 37 questions with 5 answer options instead of 4, ARC dataset size is $3530$, not $3567$.} MMLU, and CommonsenseQA.\footnote{We used validation set as test set answers are not provided.} 
The change in model performance after \SMART is shown in Table \ref{tab:smart_ranking}.
Notably, the relative ranking of models remains almost unchanged across both the original and filtered datasets. However, our methodology significantly reduces the dataset sizes, achieving up to a 68.9\% reduction for ARC, which directly correlates to a commensurate reduction in computational costs for model evaluation.

The impact of \SMART varies significantly across datasets. In ARC, 64.4\% of the examples were classified to be `easy', compared to only 27.6\% in CommonsenseQA. In MMLU, a significant 4.37\% of the dataset was found to be data contaminated, while CommonsenseQA had none. Additionally, 12.57\% of ARC and 4.56\% of MMLU examples were identified as similar. These findings highlight that different datasets might suffer from different types of low-quality examples. For MMLU, a widely accepted dataset for measuring LLM performance, we were able to filter out 43\% of the dataset while maintaining almost the same model ranking. This suggests that MMLU might not be saturated yet, and there is still room for further hill-climbing for current models.

While relative model ranking is maintained, \SMART leads to a notable decrease in inter-model agreement across all datasets, as measured by the percentage of samples yielding identical predictions among models (\autoref{fig:heatmaps}). This reduction in model agreement is beneficial, as it enables better differentiation among models based on their capabilities, and provides more headroom for improvement by decreasing their correlation.

In the following subsections, we present category-specific results for select datasets for detailed analysis, measure the correlation in model rankings before and after \SMART, and compare model accuracies with ChatBot Arena scores \cite{chiang2024chatbot}.

\begin{table}[b]
  \centering\small
  \begin{tabular}{l|c|c||c}
  \toprule
  Dataset & \multicolumn{2}{c}{Pearson's Corr.} & \multicolumn{1}{c}{Kendall's $\tau$} \\
     & \it Orig. & \it SMART & \\
    \midrule
    ARC & 0.783 & 0.845 & 0.951 \\
    MMLU & 0.764 & 0.776 & 0.978 \\
    CommonsenseQA & 0.666 & 0.660 & 0.965\\

    \bottomrule
  \end{tabular}
  \caption{Pearson correlation is high between Elo scores on ChatBot Arena and model accuracies. Kendall's Tau correlation of model ranking on original and \SMART subsets is also high.}
  \label{tab:pearson_correlation_kendall}
\end{table}

\subsection{Category-Wise Results}

Many datasets have multiple sub-categories. For instance, MMLU comprises 57 categories such as `high school mathematics', while ARC is divided into ARC-Easy and ARC-Challenge. 

Our analysis reveals significant variation in the percentage of examples removed by \SMART across different categories in MMLU.  Some categories appear to be more challenging than others. Over 60\% of examples were removed in 9 categories. Notably, `high school government and politics', `high school psychology', `marketing', and `sociology' saw a reduction in size by 73\%, 67\%, 63\%, and 62\% respectively. A qualitative look into examples of these categories indicates they contain questions with well-known answers, making them easier for models and more susceptible to be present during training and possible data contamination,
while less than 20\% were removed in $13$ categories. `Abstract algebra', `global facts', `high school physics', and `professional law' have only 4\%, 4\%, 5\%, and 12.6\% of examples removed, respectively. These subjects require complex reasoning and problem-solving, posing greater challenges for models. A detailed breakdown is shown in Appendix \ref{appsubsec:category}. 

State-of-the-art models often report accuracy on the ARC-Challenge category of ARC, as it is presumed to be more difficult \cite{meta-llama3.1, google2023gemma}. Surprisingly, our analysis shows no significant difference between ARC-Easy and ARC-Challenge after \SMART.  Both categories were significantly reduced, with 73\% of ARC-Easy and 60\% of ARC-Challenge examples removed. A high proportion, 55.1\% of ARC-Challenge were deemed easy by all models.

\subsection{Correlation in Model Rankings}\label{subsec:correlation}


To quantitatively assess the consistency of model rankings between the original datasets and their filtered versions, we employed Kendall's Tau correlation coefficient \cite{kendall1938new}. This metric ranges from $1$ (perfect positive correlation) to $-1$ (perfect negative correlation).

Our analysis of 29 models reveals high correlations between the ranking before and after \SMART. Table \ref{tab:pearson_correlation_kendall} shows the Kendall's Tau correlation coefficients for ARC, MMLU and CommonsenseQA were $0.951$, $0.978$, and $0.968$, respectively. 
The main purpose of many evaluation datasets, particularly those used in leaderboards, is to compare relative model performance. Our approach shows similar model rankings while significantly reducing dataset size (as high as $68.9\%$ for ARC) and increasing headroom.

\subsection{Correlation between ChatBot Arena scores and Model Accuracy}

ChatBot Arena \cite{chiang2024chatbot} is considered one of the most trusted current sources for assessing model performance \citep{saxon-etal-2024-benchmarks}. They use human preference to assign an Elo score \cite{elo1967proposed} for models on their leaderboard \cite{lmarena}. Some researchers consider these scores a more accurate reflection of real-world model usage than traditional benchmarks \citep{thompson-etal-2020-dataset,raji-etal-2021-everything, ott-etal-2022-mapping, saxon-etal-2024-benchmarks}. Developing high-correlation proxies for ChatBot Arena is valuable, as it could reduce the need for costly human preference collection \cite{ni2024mixeval}. 

We examine the relationship between model accuracies and ChatBot Arena Elo scores to analyze how well SMART filtered datasets reflect human preference metrics. Using Pearson correlation \cite{freedman2007statistics}, we measure the linear relationship between accuracies and Elo scores for $29$ tested models. We use Pearson correlation, because it accounts for differences in Elo scores and not just relative model rankings.

\begin{table}[t]
\centering
\resizebox{\columnwidth}{!}{
\begin{tabular}{lcccc}
\toprule
\textbf{Dataset Name} & \textbf{Original} & \textbf{SMART} & \textbf{Random} & \textbf{IRE} \\\hline
ARC & 0.783 & \textbf{0.845} & 0.774 & 0.784 \\
MMLU & 0.764 & \textbf{0.776} & 0.767 & 0.766 \\
CommonsenseQA & 0.666 & \textbf{0.660} & 0.659 & 0.658 \\
\bottomrule
\end{tabular}
}
\caption{Comparison of performance metrics across different methods. Random signifies random baseline as mentioned in Section 4.4.}
\label{tab:comparison}
\end{table}

Correlations are high across the board between ChatBot Arena Elo scores and the original versions of ARC ($r=0.783$) and MMLU ($r=0.764$); see \autoref{tab:pearson_correlation_kendall}. In fact, after \SMART the correlations \textit{increase}, suggesting that filtered ARC ($r=0.845$) and filtered MMLU ($r=0.776$) are even better proxies than the original datasets for human preference rankings. This highlights the effectiveness of \SMART in both computational efficiency and alignment to human preferences.

\subsection{Comparison with Other Approaches}

In this section we compare \SMART with other similar approaches and a random baseline. \citet{rodriguez2021evaluation} introduce a method that filters examples based on difficulty and discriminability, aiming to make leaderboards more informative. For this baseline, we used 25th percentile thresholds for difficulty and discriminability criteria to establish this baseline. We call their method IRE. We also provide analysis with a random baseline where we created a random baseline by reducing each dataset to the same size as our SMART-filtered datasets (reducing dataset size randomly to 69\%, 43\% and 34\% for ARC, MMLU and CommonsenseQA).

\begin{table}[b]
    \centering
    \small
    \begin{tabular}{lcc}
        \toprule
        Dataset Name     & \multicolumn{2}{c}{\% of Examples Filtered} \\ \hline
         & SMART & IRE \\
        \hline
        ARC & 68.9 & 28.1 \\
        MMLU & 43.0 & 31.1 \\
        CommonsenseQA & 34.3 & 27.8 \\
        \bottomrule
    \end{tabular}
    \caption{Percentage of examples filtered by SMART vs. IRE for each dataset.}
    \label{tab:filtered-removed}
\end{table}

As seen in Table \ref{tab:comparison}, SMART Filtering consistently achieves a higher correlation with human preferences than both the random baseline and IRE. Additionally, SMART Filtering consistently filters out a larger portion of the datasets compared to IRE, leading to greater efficiency in evaluation (c.f. Table \ref{tab:filtered-removed}).

\section{Robustness of our Approach}

In this section, we show that \SMART is robust to altering settings, such as (i) the number of models used for identifying easy and data-contaminated examples and (ii) choice of embedding model used to find similar examples.

\begin{table}[b]
\centering
\resizebox{\columnwidth}{!}{
\begin{tabular}{crrr}
\toprule
\# of Models & \multicolumn{3}{c}{\% of Examples Filtered} \\ \midrule
    & \it ARC & \it MMLU & \it CommonsenseQA\\ \midrule
4 & $72.9 \pm 3.1$ & $49.6 \pm 3.0$ & $41.2 \pm 4.3$\\
5 & $72.4 \pm 1.1$ & $46.1 \pm 1.6$ & $36.4 \pm 1.4$\\ 
6 & $69.7 \pm 1.1$ & $44.8 \pm 1.1$ & $34.6 \pm 1.1$\\
\bottomrule
\end{tabular}
}
\caption{Effect of varying number of models used for easy and data contamination filtering.}
\label{tab:num_models}
\end{table}

\subsection{Ablation: Number of Models for Easy and Data-Contaminated Examples} \label{sec:num_model}

In the \SMART methodology, we used 7 models for identifying easy and data-contaminated examples. Our selection was based on selecting the top 7 opensource models from different organizations. To assess the robustness of our approach, we conducted an ablation study varying the number of models used. We randomly selected subsets of 4, 5, and 6 models from the original set of 7. Specifically, we run our methodology on 10 random subsets of 4 models from a total of 35 subsets ($\binom{7}{4} = 35$), 10 random subsets of 5 models from a total of 21 subsets ($\binom{7}{5} = 21$) and all 7 subsets of 6 models ($\binom{7}{6} = 7$). The results, presented in \autoref{tab:num_models}, show that the percentage of examples filtered remains relatively stable across different model combinations for all three datasets. This shows that our methodology is largely insensitive to the number of models used, and can be effectively applied using any \textit{n} top performing open-source models, again opening up possibilities for efficiency gains.

\subsection{Embeddings for Similar Examples}

\begin{table}[t]
\centering
\begin{tabular}{lc}
\toprule
\textbf{Embedding Pair} & \textbf{Percentage Overlap}\\
\midrule
SBert \& Llama-3-8B & 89.3 \\
SBert \& Mistral-7B & 88.1 \\
\bottomrule
\end{tabular}
\vspace{-0.5em}
\caption{
Percentage of examples with extremely high semantic similarity of SentenceBert with LLM2Vec embedding, based on a distance in the embedding space smaller than a $\delta$ (see Section~\ref{sec:threshold} for more on $\delta$).
}

\label{tab:sbert_lllm2vec}
\end{table}

We use SentenceBert embeddings to capture the meaning of each example \cite{reimers-gurevych-2019-sentence}. The choice was motivated by its bi-directional attention that helps capture rich representations. Recent efforts have explored transforming decoder-only LMs into bi-directional text encoders, such as LLM2Vec \cite{behnamghader2024llm2vec}. To validate the robustness of this approach regardless of embedding model choice, we compare the model we used, SentenceBERT, with another embedding method, LLM2Vec.


Our results in \autoref{tab:sbert_lllm2vec} show a high degree of similarity between the two embedding methods, with an average overlap of 88.7\% between SentenceBERT and different LLM models. 
This suggests comparable accuracy in identifying similar examples for both SentenceBert and LLM2Vec, aligning with findings in \citet{freestone2024word} that LLM and BERT embeddings are similar. Given SentenceBert's widespread acceptance for clustering tasks and computational efficiency, we chose it for our \SMART methodology.

\subsection{Quality of Filtering}
To ensure that our method effectively filters out unwanted instances, we manually looked at 100 removed examples from the Easy and Similar examples category. For Similar examples, we found that 93 were genuinely similar, 5\% were borderline similar and 2\% were not similar. For Filtering Easy Examples, 97 examples could be easily answered by the authors or found via a simple web search. This analysis shows the effectiveness of our approach as most of the filtered out examples are unwanted for comparing model performance.

\section{Discussion}


\paragraph{Dataset Quality.} Our approach removes the easy, data contaminated, and similar examples, thus improving the quality of the resulting dataset. Our methodology can be applied at various stages, including before or after dataset release. 
\SMART is designed to be iterative and scalable; we intend to periodically reapply SMART-Filtering to existing benchmarks to identify and filter out examples that become too easy or contaminated for newer models. We suspect that, as models become more capable, what constitutes an easy example could change. 

\paragraph{Computational Efficiency.} Recent studies have explored testing models on dataset subsets while maintaining high correlation with original dataset performance \cite{varshney-etal-2022-ildae}. Our approach achieves substantial dataset size reductions, up to 68.9\% for ARC, while preserving similar model rankings as shown in Tables \ref{tab:dataset_stats} and \ref{tab:pearson_correlation_kendall}.  As dataset size correlates with computation time and evaluation costs, the 68.9\% decrease in size should be reflected in similar decreases in time and evaluation cost. Our methodology thus offers significant time and cost savings for future benchmark evaluations, aligning with the growing need for efficient model assessment techniques.

\paragraph{Improved Correlation with ChatBot Arena.} We find that \SMART can lead to better correlations with Elo scores from ChatBot Arena \cite{chiang2024chatbot}.
This suggests that model scores on \SMART subsets are more representative of real-world model performance. This finding highlights the potential for developing effective evaluation datasets that mimic real-world usage of the model without incurring substantial time and cost to capture human preferences.

\paragraph{Datasets are Not Yet Saturated.} We found that model accuracies for all tested models drop significantly on datasets after \SMART, shown in Table \ref{tab:smart_ranking}. This indicates these datasets may not be ready for retirement yet, contrary to what leaderboards might suggest, and there remains room for improvement. Relatedly, iteratively applying \SMART over time can enable more dynamic evaluations \citep{dinan-etal-2019-build, nie-etal-2020-adversarial, gehrmann-etal-2021-gem, kiela-etal-2021-dynabench, potts-etal-2021-dynasent, gehrmann-etal-2022-gemv2, margatina-etal-2023-dynamic, park-etal-2023-self, graciotti-etal-2024-latent}, decreasing the impact of saturation by wringing more utility out of existing benchmarks.


\section{Conclusion}
In this work, we proposed \SMART, a methodology for identifying a challenging and high-quality subset of any benchmark, existing or new, that can better capture the capabilities of models and rank them. To achieve this we remove easy, data contaminated, and similar examples from a dataset. \SMART achieved significant increases in computational efficiency and better correlation with human preference data than the original datasets. We anticipate our approach will be useful for improving current benchmarking practices as well as for dataset creators to find high-quality subsets before dataset release in the future.

\section{Limitations}

In this work we present a methodology that can be applied to any NLP task, However, the method we have used for identifying data contaminated examples, may not be directly applicable to non-question answering datasets. Additionally, we tried to identify and remove incorrect ground truth annotations from the dataset (see Appendix for details). However, our initial attempt did not yield satisfactory results, highlighting the need for more effective strategies to address this challenge. Consequently, if a dataset contains a significant number of annotation errors, the proportion of such examples may increase in the resulting \SMART datasets.

\bibliography{references, anthology}

\begin{thebibliography}{94}
\providecommand{\natexlab}[1]{#1}

\bibitem[{{$01.$AI} et~al.(2024){$01.$AI}, Young, Chen, Li, Huang, Zhang, Zhang, Li, Zhu, Chen, Chang, Yu, Liu, Liu, Yue, Yang, Yang, Yu, Xie, Huang, Hu, Ren, Niu, Nie, Xu, Liu, Wang, Cai, Gu, Liu, and Dai}]{ai2024yi}
{$01.$AI}, Alex Young, Bei Chen, Chao Li, Chengen Huang, Ge~Zhang, Guanwei Zhang, Heng Li, Jiangcheng Zhu, Jianqun Chen, Jing Chang, Kaidong Yu, Peng Liu, Qiang Liu, Shawn Yue, Senbin Yang, Shiming Yang, Tao Yu, Wen Xie, Wenhao Huang, Xiaohui Hu, Xiaoyi Ren, Xinyao Niu, Pengcheng Nie, Yuchi Xu, Yudong Liu, Yue Wang, Yuxuan Cai, Zhenyu Gu, Zhiyuan Liu, and Zonghong Dai. 2024.
\newblock \href {https://arxiv.org/abs/2403.04652} {Yi: Open foundation models by 01.ai}.
\newblock \emph{Preprint}, arXiv:2403.04652.

\bibitem[{Abbas et~al.(2023)Abbas, Tirumala, Simig, Ganguli, and Morcos}]{abbas2023semdedup}
Amro Abbas, Kushal Tirumala, D{\'a}niel Simig, Surya Ganguli, and Ari~S Morcos. 2023.
\newblock Semdedup: Data-efficient learning at web-scale through semantic deduplication.
\newblock \emph{Multimodal Representation Learning (MRL): Perks and Pitfalls}.

\bibitem[{Abdin et~al.(2024)Abdin, Jacobs, Awan, Aneja, Awadallah, Awadalla, Bach, Bahree, Bakhtiari, Behl et~al.}]{abdin2024phi}
Marah Abdin, Sam~Ade Jacobs, Ammar~Ahmad Awan, Jyoti Aneja, Ahmed Awadallah, Hany Awadalla, Nguyen Bach, Amit Bahree, Arash Bakhtiari, Harkirat Behl, et~al. 2024.
\newblock Phi-3 technical report: A highly capable language model locally on your phone.
\newblock \emph{arXiv preprint arXiv:2404.14219}.

\bibitem[{Alzahrani et~al.(2024)Alzahrani, Alyahya, Alnumay, AlRashed, Alsubaie, Almushayqih, Mirza, Alotaibi, Al-Twairesh, Alowisheq, Bari, and Khan}]{alzahrani-etal-2024-benchmarks}
Norah Alzahrani, Hisham Alyahya, Yazeed Alnumay, Sultan AlRashed, Shaykhah Alsubaie, Yousef Almushayqih, Faisal Mirza, Nouf Alotaibi, Nora Al-Twairesh, Areeb Alowisheq, M~Saiful Bari, and Haidar Khan. 2024.
\newblock \href {https://aclanthology.org/2024.acl-long.744} {When benchmarks are targets: Revealing the sensitivity of large language model leaderboards}.
\newblock In \emph{Proceedings of the 62nd Annual Meeting of the Association for Computational Linguistics (Volume 1: Long Papers)}, pages 13787--13805, Bangkok, Thailand. Association for Computational Linguistics.

\bibitem[{Balepur et~al.(2024)Balepur, Ravichander, and Rudinger}]{balepur-etal-2024-artifacts}
Nishant Balepur, Abhilasha Ravichander, and Rachel Rudinger. 2024.
\newblock \href {https://aclanthology.org/2024.acl-long.555} {Artifacts or abduction: How do {LLM}s answer multiple-choice questions without the question?}
\newblock In \emph{Proceedings of the 62nd Annual Meeting of the Association for Computational Linguistics (Volume 1: Long Papers)}, pages 10308--10330, Bangkok, Thailand. Association for Computational Linguistics.

\bibitem[{Balepur and Rudinger(2024)}]{balepur-rudinger-2024-large}
Nishant Balepur and Rachel Rudinger. 2024.
\newblock \href {https://aclanthology.org/2024.knowllm-1.2} {Is your large language model knowledgeable or a choices-only cheater?}
\newblock In \emph{Proceedings of the 1st Workshop on Towards Knowledgeable Language Models (KnowLLM 2024)}, pages 15--26, Bangkok, Thailand. Association for Computational Linguistics.

\bibitem[{BehnamGhader et~al.(2024)BehnamGhader, Adlakha, Mosbach, Bahdanau, Chapados, and Reddy}]{behnamghader2024llm2vec}
Parishad BehnamGhader, Vaibhav Adlakha, Marius Mosbach, Dzmitry Bahdanau, Nicolas Chapados, and Siva Reddy. 2024.
\newblock Llm2vec: Large language models are secretly powerful text encoders.
\newblock \emph{arXiv preprint arXiv:2404.05961}.

\bibitem[{Bras et~al.(2020)Bras, Swayamdipta, Bhagavatula, Zellers, Peters, Sabharwal, and Choi}]{pmlr-v119-bras20a}
Ronan~Le Bras, Swabha Swayamdipta, Chandra Bhagavatula, Rowan Zellers, Matthew Peters, Ashish Sabharwal, and Yejin Choi. 2020.
\newblock \href {https://proceedings.mlr.press/v119/bras20a.html} {Adversarial filters of dataset biases}.
\newblock In \emph{Proceedings of the 37th International Conference on Machine Learning}, volume 119 of \emph{Proceedings of Machine Learning Research}, pages 1078--1088. PMLR.

\bibitem[{Chiang et~al.(2024)Chiang, Zheng, Sheng, Angelopoulos, Li, Li, Zhu, Zhang, Jordan, Gonzalez, and Stoica}]{chiang2024chatbot}
Wei-Lin Chiang, Lianmin Zheng, Ying Sheng, Anastasios~Nikolas Angelopoulos, Tianle Li, Dacheng Li, Banghua Zhu, Hao Zhang, Michael Jordan, Joseph~E. Gonzalez, and Ion Stoica. 2024.
\newblock \href {https://openreview.net/forum?id=3MW8GKNyzI} {Chatbot arena: An open platform for evaluating {LLM}s by human preference}.
\newblock In \emph{Forty-first International Conference on Machine Learning}.

\bibitem[{{Chiang et al.}([2024])}]{lmarena}
{Chiang et al.} [2024].
\newblock {ChatBot Arena}.
\newblock \url{https://lmarena.ai/?leaderboard}.

\bibitem[{Clark et~al.(2018)Clark, Cowhey, Etzioni, Khot, Sabharwal, Schoenick, and Tafjord}]{clark2018think}
Peter Clark, Isaac Cowhey, Oren Etzioni, Tushar Khot, Ashish Sabharwal, Carissa Schoenick, and Oyvind Tafjord. 2018.
\newblock Think you have solved question answering? try arc, the ai2 reasoning challenge.
\newblock \emph{arXiv preprint arXiv:1803.05457}.

\bibitem[{Cobbe et~al.(2021)Cobbe, Kosaraju, Bavarian, Chen, Jun, Kaiser, Plappert, Tworek, Hilton, Nakano et~al.}]{cobbe2021training}
Karl Cobbe, Vineet Kosaraju, Mohammad Bavarian, Mark Chen, Heewoo Jun, Lukasz Kaiser, Matthias Plappert, Jerry Tworek, Jacob Hilton, Reiichiro Nakano, et~al. 2021.
\newblock Training verifiers to solve math word problems.
\newblock \emph{arXiv preprint arXiv:2110.14168}.

\bibitem[{{Databricks}(2024)}]{databricks2023dbrx}
{Databricks}. 2024.
\newblock \href {https://www.databricks.com/blog/introducing-dbrx-new-state-art-open-llm} {{Introducing DBRx: A New State-of-the-Art Open LLM}}.

\bibitem[{Deng et~al.(2023)Deng, Zhao, Tang, Gerstein, and Cohan}]{deng2023benchmark}
Chunyuan Deng, Yilun Zhao, Xiangru Tang, Mark Gerstein, and Arman Cohan. 2023.
\newblock Benchmark probing: Investigating data leakage in large language models.
\newblock In \emph{NeurIPS 2023 Workshop on Backdoors in Deep Learning-The Good, the Bad, and the Ugly}.

\bibitem[{Dinan et~al.(2019)Dinan, Humeau, Chintagunta, and Weston}]{dinan-etal-2019-build}
Emily Dinan, Samuel Humeau, Bharath Chintagunta, and Jason Weston. 2019.
\newblock \href {https://doi.org/10.18653/v1/D19-1461} {Build it break it fix it for dialogue safety: Robustness from adversarial human attack}.
\newblock In \emph{Proceedings of the 2019 Conference on Empirical Methods in Natural Language Processing and the 9th International Joint Conference on Natural Language Processing (EMNLP-IJCNLP)}, pages 4537--4546, Hong Kong, China. Association for Computational Linguistics.

\bibitem[{Duan et~al.(2024)Duan, Suri, Mireshghallah, Min, Shi, Zettlemoyer, Tsvetkov, Choi, Evans, and Hajishirzi}]{duan2024membership}
Michael Duan, Anshuman Suri, Niloofar Mireshghallah, Sewon Min, Weijia Shi, Luke Zettlemoyer, Yulia Tsvetkov, Yejin Choi, David Evans, and Hannaneh Hajishirzi. 2024.
\newblock Do membership inference attacks work on large language models?
\newblock \emph{arXiv preprint arXiv:2402.07841}.

\bibitem[{Dugan et~al.(2023)Dugan, Ippolito, Kirubarajan, Shi, and Callison-Burch}]{dugan-etal-2023-real}
Liam Dugan, Daphne Ippolito, Arun Kirubarajan, Sherry Shi, and Chris Callison-Burch. 2023.
\newblock Real or fake text?: Investigating human ability to detect boundaries between human-written and machine-generated text.
\newblock In \emph{Proceedings of the AAAI Conference on Artificial Intelligence}, volume~37, pages 12763--12771.

\bibitem[{Elangovan et~al.(2021)Elangovan, He, and Verspoor}]{elangovan-etal-2021-memorization}
Aparna Elangovan, Jiayuan He, and Karin Verspoor. 2021.
\newblock \href {https://doi.org/10.18653/v1/2021.eacl-main.113} {Memorization vs. generalization : Quantifying data leakage in {NLP} performance evaluation}.
\newblock In \emph{Proceedings of the 16th Conference of the European Chapter of the Association for Computational Linguistics: Main Volume}, pages 1325--1335, Online. Association for Computational Linguistics.

\bibitem[{Elazar et~al.(2024)Elazar, Bhagia, Magnusson, Ravichander, Schwenk, Suhr, Walsh, Groeneveld, Soldaini, Singh, Hajishirzi, Smith, and Dodge}]{elazar2024whats}
Yanai Elazar, Akshita Bhagia, Ian~Helgi Magnusson, Abhilasha Ravichander, Dustin Schwenk, Alane Suhr, Evan~Pete Walsh, Dirk Groeneveld, Luca Soldaini, Sameer Singh, Hannaneh Hajishirzi, Noah~A. Smith, and Jesse Dodge. 2024.
\newblock \href {https://openreview.net/forum?id=RvfPnOkPV4} {What's in my big data?}
\newblock In \emph{The Twelfth International Conference on Learning Representations}.

\bibitem[{Elo(1967)}]{elo1967proposed}
Arpad~E Elo. 1967.
\newblock The proposed uscf rating system, its development, theory, and applications.
\newblock \emph{Chess life}, 22(8):242--247.

\bibitem[{Freedman et~al.(2007)Freedman, Pisani, and Purves}]{freedman2007statistics}
David Freedman, Robert Pisani, and Roger Purves. 2007.
\newblock Statistics (international student edition).
\newblock \emph{Pisani, R. Purves, 4th edn. WW Norton \& Company, New York}.

\bibitem[{Freestone and Santu(2024)}]{freestone2024word}
Matthew Freestone and Shubhra Kanti~Karmaker Santu. 2024.
\newblock Word embeddings revisited: Do llms offer something new?
\newblock \emph{arXiv preprint arXiv:2402.11094}.

\bibitem[{Gehrmann et~al.(2021)Gehrmann, Adewumi, Aggarwal, Ammanamanchi, Aremu, Bosselut, Chandu, Clinciu, Das, Dhole, Du, Durmus, Du{\v{s}}ek, Emezue, Gangal, Garbacea, Hashimoto, Hou, Jernite, Jhamtani, Ji, Jolly, Kale, Kumar, Ladhak, Madaan, Maddela, Mahajan, Mahamood, Majumder, Martins, McMillan-Major, Mille, van Miltenburg, Nadeem, Narayan, Nikolaev, Niyongabo~Rubungo, Osei, Parikh, Perez-Beltrachini, Rao, Raunak, Rodriguez, Santhanam, Sedoc, Sellam, Shaikh, Shimorina, Sobrevilla~Cabezudo, Strobelt, Subramani, Xu, Yang, Yerukola, and Zhou}]{gehrmann-etal-2021-gem}
Sebastian Gehrmann, Tosin Adewumi, Karmanya Aggarwal, Pawan~Sasanka Ammanamanchi, Anuoluwapo Aremu, Antoine Bosselut, Khyathi~Raghavi Chandu, Miruna-Adriana Clinciu, Dipanjan Das, Kaustubh Dhole, Wanyu Du, Esin Durmus, Ond{\v{r}}ej Du{\v{s}}ek, Chris~Chinenye Emezue, Varun Gangal, Cristina Garbacea, Tatsunori Hashimoto, Yufang Hou, Yacine Jernite, Harsh Jhamtani, Yangfeng Ji, Shailza Jolly, Mihir Kale, Dhruv Kumar, Faisal Ladhak, Aman Madaan, Mounica Maddela, Khyati Mahajan, Saad Mahamood, Bodhisattwa~Prasad Majumder, Pedro~Henrique Martins, Angelina McMillan-Major, Simon Mille, Emiel van Miltenburg, Moin Nadeem, Shashi Narayan, Vitaly Nikolaev, Andre Niyongabo~Rubungo, Salomey Osei, Ankur Parikh, Laura Perez-Beltrachini, Niranjan~Ramesh Rao, Vikas Raunak, Juan~Diego Rodriguez, Sashank Santhanam, Jo{\~a}o Sedoc, Thibault Sellam, Samira Shaikh, Anastasia Shimorina, Marco~Antonio Sobrevilla~Cabezudo, Hendrik Strobelt, Nishant Subramani, Wei Xu, Diyi Yang, Akhila Yerukola, and Jiawei Zhou. 2021.
\newblock \href {https://doi.org/10.18653/v1/2021.gem-1.10} {The {GEM} benchmark: Natural language generation, its evaluation and metrics}.
\newblock In \emph{Proceedings of the 1st Workshop on Natural Language Generation, Evaluation, and Metrics (GEM 2021)}, pages 96--120, Online. Association for Computational Linguistics.

\bibitem[{Gehrmann et~al.(2022)Gehrmann, Bhattacharjee, Mahendiran, Wang, Papangelis, Madaan, Mcmillan-major, Shvets, Upadhyay, Bohnet, Yao, Wilie, Bhagavatula, You, Thomson, Garbacea, Wang, Deutsch, Xiong, Jin, Gkatzia, Radev, Clark, Durmus, Ladhak, Ginter, Winata, Strobelt, Hayashi, Novikova, Kanerva, Chim, Zhou, Clive, Maynez, Sedoc, Juraska, Dhole, Chandu, Beltrachini, Ribeiro, Tunstall, Zhang, Pushkarna, Creutz, White, Kale, Eddine, Daheim, Subramani, Dusek, Liang, Ammanamanchi, Zhu, Puduppully, Kriz, Shahriyar, Cardenas, Mahamood, Osei, Cahyawijaya, {\v{S}}tajner, Montella, Jolly, Mille, Hasan, Shen, Adewumi, Raunak, Raheja, Nikolaev, Tsai, Jernite, Xu, Sang, Liu, and Hou}]{gehrmann-etal-2022-gemv2}
Sebastian Gehrmann, Abhik Bhattacharjee, Abinaya Mahendiran, Alex Wang, Alexandros Papangelis, Aman Madaan, Angelina Mcmillan-major, Anna Shvets, Ashish Upadhyay, Bernd Bohnet, Bingsheng Yao, Bryan Wilie, Chandra Bhagavatula, Chaobin You, Craig Thomson, Cristina Garbacea, Dakuo Wang, Daniel Deutsch, Deyi Xiong, Di~Jin, Dimitra Gkatzia, Dragomir Radev, Elizabeth Clark, Esin Durmus, Faisal Ladhak, Filip Ginter, Genta~Indra Winata, Hendrik Strobelt, Hiroaki Hayashi, Jekaterina Novikova, Jenna Kanerva, Jenny Chim, Jiawei Zhou, Jordan Clive, Joshua Maynez, Jo{\~a}o Sedoc, Juraj Juraska, Kaustubh Dhole, Khyathi~Raghavi Chandu, Laura~Perez Beltrachini, Leonardo F .~R. Ribeiro, Lewis Tunstall, Li~Zhang, Mahim Pushkarna, Mathias Creutz, Michael White, Mihir~Sanjay Kale, Moussa~Kamal Eddine, Nico Daheim, Nishant Subramani, Ondrej Dusek, Paul~Pu Liang, Pawan~Sasanka Ammanamanchi, Qi~Zhu, Ratish Puduppully, Reno Kriz, Rifat Shahriyar, Ronald Cardenas, Saad Mahamood, Salomey Osei, Samuel Cahyawijaya, Sanja {\v{S}}tajner,
  Sebastien Montella, Shailza Jolly, Simon Mille, Tahmid Hasan, Tianhao Shen, Tosin Adewumi, Vikas Raunak, Vipul Raheja, Vitaly Nikolaev, Vivian Tsai, Yacine Jernite, Ying Xu, Yisi Sang, Yixin Liu, and Yufang Hou. 2022.
\newblock \href {https://doi.org/10.18653/v1/2022.emnlp-demos.27} {{GEM}v2: Multilingual {NLG} benchmarking in a single line of code}.
\newblock In \emph{Proceedings of the 2022 Conference on Empirical Methods in Natural Language Processing: System Demonstrations}, pages 266--281, Abu Dhabi, UAE. Association for Computational Linguistics.

\bibitem[{Gema et~al.(2024)Gema, Leang, Hong, Devoto, Mancino, Saxena, He, Zhao, Du, Madani et~al.}]{gema-etal-2024-we}
Aryo~Pradipta Gema, Joshua Ong~Jun Leang, Giwon Hong, Alessio Devoto, Alberto Carlo~Maria Mancino, Rohit Saxena, Xuanli He, Yu~Zhao, Xiaotang Du, Mohammad Reza~Ghasemi Madani, et~al. 2024.
\newblock Are we done with mmlu?
\newblock \emph{arXiv preprint arXiv:2406.04127}.

\bibitem[{Google(2024)}]{google2023gemma}
Google. 2024.
\newblock \href {https://blog.google/technology/developers/google-gemma-2/} {{Google Gemma 2: A New AI Model}}.

\bibitem[{Graciotti et~al.(2024)Graciotti, Presutti, and Tripodi}]{graciotti-etal-2024-latent}
Arianna Graciotti, Valentina Presutti, and Rocco Tripodi. 2024.
\newblock \href {https://aclanthology.org/2024.lrec-main.888} {Latent vs explicit knowledge representation: How {C}hat{GPT} answers questions about low-frequency entities}.
\newblock In \emph{Proceedings of the 2024 Joint International Conference on Computational Linguistics, Language Resources and Evaluation (LREC-COLING 2024)}, pages 10172--10185, Torino, Italia. ELRA and ICCL.

\bibitem[{Gupta et~al.(2024)Gupta, Pantoja, Ross, Williams, and Ung}]{gupta2024changing}
Vipul Gupta, David Pantoja, Candace Ross, Adina Williams, and Megan Ung. 2024.
\newblock Changing answer order can decrease {MMLU} accuracy.
\newblock \emph{arXiv preprint arXiv:2406.19470}.

\bibitem[{Gupta et~al.(2023)Gupta, Venkit, Lauren{\c{c}}on, Wilson, and Passonneau}]{gupta2023calm}
Vipul Gupta, Pranav~Narayanan Venkit, Hugo Lauren{\c{c}}on, Shomir Wilson, and Rebecca~J Passonneau. 2023.
\newblock Calm: A multi-task benchmark for comprehensive assessment of language model bias.
\newblock \emph{arXiv preprint arXiv:2308.12539}.

\bibitem[{Gururangan et~al.(2018)Gururangan, Swayamdipta, Levy, Schwartz, Bowman, and Smith}]{gururangan-etal-2018-annotation}
Suchin Gururangan, Swabha Swayamdipta, Omer Levy, Roy Schwartz, Samuel Bowman, and Noah~A. Smith. 2018.
\newblock \href {https://doi.org/10.18653/v1/N18-2017} {Annotation artifacts in natural language inference data}.
\newblock In \emph{Proceedings of the 2018 Conference of the North {A}merican Chapter of the Association for Computational Linguistics: Human Language Technologies, Volume 2 (Short Papers)}, pages 107--112, New Orleans, Louisiana. Association for Computational Linguistics.

\bibitem[{Hendrycks et~al.(2020)Hendrycks, Burns, Basart, Critch, Li, Song, and Steinhardt}]{hendrycks-etal-2020-aligning}
Dan Hendrycks, Collin Burns, Steven Basart, Andrew Critch, Jerry Li, Dawn Song, and Jacob Steinhardt. 2020.
\newblock Aligning ai with shared human values.
\newblock \emph{arXiv preprint arXiv:2008.02275}.

\bibitem[{Hendrycks et~al.(2021{\natexlab{a}})Hendrycks, Burns, Basart, Zou, Mazeika, Song, and Steinhardt}]{hendrycks2021measuring}
Dan Hendrycks, Collin Burns, Steven Basart, Andy Zou, Mantas Mazeika, Dawn Song, and Jacob Steinhardt. 2021{\natexlab{a}}.
\newblock \href {https://arxiv.org/abs/2009.03300} {Measuring massive multitask language understanding}.
\newblock \emph{Preprint}, arXiv:2009.03300.

\bibitem[{Hendrycks et~al.(2021{\natexlab{b}})Hendrycks, Burns, Kadavath, Arora, Basart, Tang, Song, and Steinhardt}]{hendrycks2021math}
Dan Hendrycks, Collin Burns, Saurav Kadavath, Akul Arora, Steven Basart, Eric Tang, Dawn Song, and Jacob Steinhardt. 2021{\natexlab{b}}.
\newblock Measuring mathematical problem solving with the math dataset.
\newblock \emph{arXiv preprint arXiv:2103.03874}.

\bibitem[{Herde et~al.(2021)Herde, Huseljic, Sick, and Calma}]{herde2021survey}
Marek Herde, Denis Huseljic, Bernhard Sick, and Adrian Calma. 2021.
\newblock A survey on cost types, interaction schemes, and annotator performance models in selection algorithms for active learning in classification.
\newblock \emph{IEEE Access}, 9:166970--166989.

\bibitem[{HuggingFace([2024])}]{open-llm-leaderboard}
HuggingFace. [2024].
\newblock {Open LLM Leaderboard}.
\newblock \url{https://huggingface.co/spaces/open-llm-leaderboard/open_llm_leaderboard}.

\bibitem[{Jawahar et~al.(2020)Jawahar, Abdul-Mageed, and Lakshmanan}]{jawahar-etal-2020-automatic}
Ganesh Jawahar, Muhammad Abdul-Mageed, and Laks Lakshmanan, V.S. 2020.
\newblock \href {https://doi.org/10.18653/v1/2020.coling-main.208} {Automatic detection of machine generated text: A critical survey}.
\newblock In \emph{Proceedings of the 28th International Conference on Computational Linguistics}, pages 2296--2309, Barcelona, Spain (Online). International Committee on Computational Linguistics.

\bibitem[{Jiang et~al.(2024)Jiang, Liu, Zhong, Schaeffer, Ouyang, Han, and Koyejo}]{jiang2024investigating}
Minhao Jiang, Ken~Ziyu Liu, Ming Zhong, Rylan Schaeffer, Siru Ouyang, Jiawei Han, and Sanmi Koyejo. 2024.
\newblock Investigating data contamination for pre-training language models.
\newblock \emph{arXiv preprint arXiv:2401.06059}.

\bibitem[{Keegan(2024)}]{keeganblog}
Jon Keegan. 2024.
\newblock \href {https://themarkup.org/artificial-intelligence/2024/07/17/everyone-is-judging-ai-by-these-tests-but-experts-say-theyre-close-to-meaningless} {Everyone is judging ai by these tests, but experts say they're close to meaningless}.
\newblock Accessed: [Insert Date].

\bibitem[{Kendall(1938)}]{kendall1938new}
Maurice~G Kendall. 1938.
\newblock A new measure of rank correlation.
\newblock \emph{Biometrika}, 30(1-2):81--93.

\bibitem[{Khattab and Zaharia(2020)}]{khattab2020colbert}
Omar Khattab and Matei Zaharia. 2020.
\newblock Colbert: Efficient and effective passage search via contextualized late interaction over bert.
\newblock In \emph{Proceedings of the 43rd International ACM SIGIR conference on research and development in Information Retrieval}, pages 39--48.

\bibitem[{Kiela et~al.(2021)Kiela, Bartolo, Nie, Kaushik, Geiger, Wu, Vidgen, Prasad, Singh, Ringshia, Ma, Thrush, Riedel, Waseem, Stenetorp, Jia, Bansal, Potts, and Williams}]{kiela-etal-2021-dynabench}
Douwe Kiela, Max Bartolo, Yixin Nie, Divyansh Kaushik, Atticus Geiger, Zhengxuan Wu, Bertie Vidgen, Grusha Prasad, Amanpreet Singh, Pratik Ringshia, Zhiyi Ma, Tristan Thrush, Sebastian Riedel, Zeerak Waseem, Pontus Stenetorp, Robin Jia, Mohit Bansal, Christopher Potts, and Adina Williams. 2021.
\newblock \href {https://doi.org/10.18653/v1/2021.naacl-main.324} {Dynabench: Rethinking benchmarking in {NLP}}.
\newblock In \emph{Proceedings of the 2021 Conference of the North American Chapter of the Association for Computational Linguistics: Human Language Technologies}, pages 4110--4124, Online. Association for Computational Linguistics.

\bibitem[{Laskar et~al.(2023)Laskar, Bari, Rahman, Bhuiyan, Joty, and Huang}]{laskar-etal-2023-systematic}
Md~Tahmid~Rahman Laskar, M~Saiful Bari, Mizanur Rahman, Md~Amran~Hossen Bhuiyan, Shafiq Joty, and Jimmy Huang. 2023.
\newblock \href {https://doi.org/10.18653/v1/2023.findings-acl.29} {A systematic study and comprehensive evaluation of {C}hat{GPT} on benchmark datasets}.
\newblock In \emph{Findings of the Association for Computational Linguistics: ACL 2023}, pages 431--469, Toronto, Canada. Association for Computational Linguistics.

\bibitem[{Lee et~al.(2022)Lee, Ippolito, Nystrom, Zhang, Eck, Callison-Burch, and Carlini}]{lee-etal-2022-deduplicating}
Katherine Lee, Daphne Ippolito, Andrew Nystrom, Chiyuan Zhang, Douglas Eck, Chris Callison-Burch, and Nicholas Carlini. 2022.
\newblock \href {https://doi.org/10.18653/v1/2022.acl-long.577} {Deduplicating training data makes language models better}.
\newblock In \emph{Proceedings of the 60th Annual Meeting of the Association for Computational Linguistics (Volume 1: Long Papers)}, pages 8424--8445, Dublin, Ireland. Association for Computational Linguistics.

\bibitem[{Li and Gao(2024)}]{li2024anchored}
Ruizhe Li and Yanjun Gao. 2024.
\newblock Anchored answers: Unravelling positional bias in gpt-2's multiple-choice questions.
\newblock \emph{arXiv preprint arXiv:2405.03205}.

\bibitem[{Longpre et~al.(2024)Longpre, Mahari, Lee, Lund, Oderinwale, Brannon, Saxena, Obeng-Marnu, South, Hunter et~al.}]{longpre2024consent}
Shayne Longpre, Robert Mahari, Ariel Lee, Campbell Lund, Hamidah Oderinwale, William Brannon, Nayan Saxena, Naana Obeng-Marnu, Tobin South, Cole Hunter, et~al. 2024.
\newblock Consent in crisis: The rapid decline of the ai data commons.
\newblock \emph{arXiv preprint arXiv:2407.14933}.

\bibitem[{Ma et~al.(2024)Ma, Wang, Yang, Wei, and Lin}]{ma2024fine}
Xueguang Ma, Liang Wang, Nan Yang, Furu Wei, and Jimmy Lin. 2024.
\newblock Fine-tuning llama for multi-stage text retrieval.
\newblock In \emph{Proceedings of the 47th International ACM SIGIR Conference on Research and Development in Information Retrieval}, pages 2421--2425.

\bibitem[{Magar and Schwartz(2022)}]{magar-schwartz-2022-data}
Inbal Magar and Roy Schwartz. 2022.
\newblock \href {https://doi.org/10.18653/v1/2022.acl-short.18} {Data contamination: From memorization to exploitation}.
\newblock In \emph{Proceedings of the 60th Annual Meeting of the Association for Computational Linguistics (Volume 2: Short Papers)}, pages 157--165, Dublin, Ireland. Association for Computational Linguistics.

\bibitem[{Margatina et~al.(2023)Margatina, Wang, Vyas, Anna~John, Benajiba, and Ballesteros}]{margatina-etal-2023-dynamic}
Katerina Margatina, Shuai Wang, Yogarshi Vyas, Neha Anna~John, Yassine Benajiba, and Miguel Ballesteros. 2023.
\newblock \href {https://doi.org/10.18653/v1/2023.eacl-main.211} {Dynamic benchmarking of masked language models on temporal concept drift with multiple views}.
\newblock In \emph{Proceedings of the 17th Conference of the European Chapter of the Association for Computational Linguistics}, pages 2881--2898, Dubrovnik, Croatia. Association for Computational Linguistics.

\bibitem[{Matatov et~al.(2022)Matatov, Naaman, and Amir}]{matatov2022dataset}
Hana Matatov, Mor Naaman, and Ofra Amir. 2022.
\newblock Dataset and case studies for visual near-duplicates detection in the context of social media.
\newblock \emph{arXiv preprint arXiv:2203.07167}.

\bibitem[{McIlroy-Young et~al.(2024)McIlroy-Young, Brown, Olson, Zhang, and Dwork}]{mcilroyyoung2024setbasedpromptingprovablysolving}
Reid McIlroy-Young, Katrina Brown, Conlan Olson, Linjun Zhang, and Cynthia Dwork. 2024.
\newblock \href {https://arxiv.org/abs/2406.06581} {Set-based prompting: Provably solving the language model order dependency problem}.
\newblock \emph{Preprint}, arXiv:2406.06581.

\bibitem[{{Meta AI}(2024)}]{meta-llama3.1}
{Meta AI}. 2024.
\newblock {Meta LLaMA 3.1}.
\newblock \url{https://ai.meta.com/blog/meta-llama-3-1/}.
\newblock Accessed: [Insert Date].

\bibitem[{Mishra and Sachdeva(2020)}]{mishra-sachdeva-2020-need}
Swaroop Mishra and Bhavdeep~Singh Sachdeva. 2020.
\newblock \href {https://doi.org/10.18653/v1/2020.sustainlp-1.23} {Do we need to create big datasets to learn a task?}
\newblock In \emph{Proceedings of SustaiNLP: Workshop on Simple and Efficient Natural Language Processing}, pages 169--173, Online. Association for Computational Linguistics.

\bibitem[{{Mistral AI}(2023)}]{mistral2023mixtral}
{Mistral AI}. 2023.
\newblock \href {https://mistral.ai/news/mixtral-of-experts/} {{Mixtral of Experts}}.
\newblock Section on Mixtral Architecture. Accessed: [Insert Date].

\bibitem[{Neelakantan et~al.(2022)Neelakantan, Xu, Puri, Radford, Han, Tworek, Yuan, Tezak, Kim, Hallacy et~al.}]{neelakantan2022text}
Arvind Neelakantan, Tao Xu, Raul Puri, Alec Radford, Jesse~Michael Han, Jerry Tworek, Qiming Yuan, Nikolas Tezak, Jong~Wook Kim, Chris Hallacy, et~al. 2022.
\newblock Text and code embeddings by contrastive pre-training.
\newblock \emph{arXiv preprint arXiv:2201.10005}.

\bibitem[{Ni et~al.(2024)Ni, Xue, Yue, Deng, Shah, Jain, Neubig, and You}]{ni2024mixeval}
Jinjie Ni, Fuzhao Xue, Xiang Yue, Yuntian Deng, Mahir Shah, Kabir Jain, Graham Neubig, and Yang You. 2024.
\newblock Mixeval: Deriving wisdom of the crowd from llm benchmark mixtures.
\newblock \emph{arXiv preprint arXiv:2406.06565}.

\bibitem[{Nie et~al.(2020)Nie, Williams, Dinan, Bansal, Weston, and Kiela}]{nie-etal-2020-adversarial}
Yixin Nie, Adina Williams, Emily Dinan, Mohit Bansal, Jason Weston, and Douwe Kiela. 2020.
\newblock \href {https://doi.org/10.18653/v1/2020.acl-main.441} {Adversarial {NLI}: A new benchmark for natural language understanding}.
\newblock In \emph{Proceedings of the 58th Annual Meeting of the Association for Computational Linguistics}, pages 4885--4901, Online. Association for Computational Linguistics.

\bibitem[{Ott et~al.(2022)Ott, Barbosa-Silva, Blagec, Brauner, and Samwald}]{ott-etal-2022-mapping}
Simon Ott, Adriano Barbosa-Silva, Kathrin Blagec, Jan Brauner, and Matthias Samwald. 2022.
\newblock Mapping global dynamics of benchmark creation and saturation in artificial intelligence.
\newblock \emph{Nature Communications}, 13(1):6793.

\bibitem[{Park et~al.(2023)Park, Moon, Lee, Seo, Eo, and Lim}]{park-etal-2023-self}
Chanjun Park, Hyeonseok Moon, Seolhwa Lee, Jaehyung Seo, Sugyeong Eo, and Heuiseok Lim. 2023.
\newblock Self-improving-leaderboard (sil): A call for real-world centric natural language processing leaderboards.
\newblock \emph{arXiv preprint arXiv:2303.10888}.

\bibitem[{Pezeshkpour and Hruschka(2024)}]{pezeshkpour-hruschka-2024-large}
Pouya Pezeshkpour and Estevam Hruschka. 2024.
\newblock \href {https://doi.org/10.18653/v1/2024.findings-naacl.130} {Large language models sensitivity to the order of options in multiple-choice questions}.
\newblock In \emph{Findings of the Association for Computational Linguistics: NAACL 2024}, pages 2006--2017, Mexico City, Mexico. Association for Computational Linguistics.

\bibitem[{Phang et~al.(2022)Phang, Chen, Huang, and Bowman}]{phang-etal-2022-adversarially}
Jason Phang, Angelica Chen, William Huang, and Samuel~R. Bowman. 2022.
\newblock \href {https://doi.org/10.18653/v1/2022.dadc-1.8} {Adversarially constructed evaluation sets are more challenging, but may not be fair}.
\newblock In \emph{Proceedings of the First Workshop on Dynamic Adversarial Data Collection}, pages 62--62, Seattle, WA. Association for Computational Linguistics.

\bibitem[{Potts et~al.(2021)Potts, Wu, Geiger, and Kiela}]{potts-etal-2021-dynasent}
Christopher Potts, Zhengxuan Wu, Atticus Geiger, and Douwe Kiela. 2021.
\newblock \href {https://doi.org/10.18653/v1/2021.acl-long.186} {{D}yna{S}ent: A dynamic benchmark for sentiment analysis}.
\newblock In \emph{Proceedings of the 59th Annual Meeting of the Association for Computational Linguistics and the 11th International Joint Conference on Natural Language Processing (Volume 1: Long Papers)}, pages 2388--2404, Online. Association for Computational Linguistics.

\bibitem[{Raj et~al.(2023)Raj, Gupta, Rosati, and Majumdar}]{raj2023semantic}
Harsh Raj, Vipul Gupta, Domenic Rosati, and Subhabrata Majumdar. 2023.
\newblock Semantic consistency for assuring reliability of large language models.
\newblock \emph{arXiv preprint arXiv:2308.09138}.

\bibitem[{Raji et~al.(2021)Raji, Denton, Bender, Hanna, and Paullada}]{raji-etal-2021-everything}
Deborah Raji, Emily Denton, Emily~M. Bender, Alex Hanna, and Amandalynne Paullada. 2021.
\newblock \href {https://datasets-benchmarks-proceedings.neurips.cc/paper_files/paper/2021/file/084b6fbb10729ed4da8c3d3f5a3ae7c9-Paper-round2.pdf} {Ai and the everything in the whole wide world benchmark}.
\newblock In \emph{Proceedings of the Neural Information Processing Systems Track on Datasets and Benchmarks}, volume~1.

\bibitem[{Ravaut et~al.(2024)Ravaut, Ding, Jiao, Chen, Li, Zhao, Qin, Xiong, and Joty}]{ravaut2024much}
Mathieu Ravaut, Bosheng Ding, Fangkai Jiao, Hailin Chen, Xingxuan Li, Ruochen Zhao, Chengwei Qin, Caiming Xiong, and Shafiq Joty. 2024.
\newblock How much are llms contaminated? a comprehensive survey and the llmsanitize library.
\newblock \emph{arXiv preprint arXiv:2404.00699}.

\bibitem[{Reif and Schwartz(2024)}]{reif-schwartz-2024-beyond}
Yuval Reif and Roy Schwartz. 2024.
\newblock \href {https://doi.org/10.18653/v1/2024.naacl-long.378} {Beyond performance: Quantifying and mitigating label bias in {LLM}s}.
\newblock In \emph{Proceedings of the 2024 Conference of the North American Chapter of the Association for Computational Linguistics: Human Language Technologies (Volume 1: Long Papers)}, pages 6784--6798, Mexico City, Mexico. Association for Computational Linguistics.

\bibitem[{Reimers and Gurevych(2019)}]{reimers-gurevych-2019-sentence}
Nils Reimers and Iryna Gurevych. 2019.
\newblock \href {https://doi.org/10.18653/v1/D19-1410} {Sentence-{BERT}: Sentence embeddings using {S}iamese {BERT}-networks}.
\newblock In \emph{Proceedings of the 2019 Conference on Empirical Methods in Natural Language Processing and the 9th International Joint Conference on Natural Language Processing (EMNLP-IJCNLP)}, pages 3982--3992, Hong Kong, China. Association for Computational Linguistics.

\bibitem[{Rein et~al.(2023)Rein, Hou, Stickland, Petty, Pang, Dirani, Michael, and Bowman}]{rein2023gpqa}
David Rein, Betty~Li Hou, Asa~Cooper Stickland, Jackson Petty, Richard~Yuanzhe Pang, Julien Dirani, Julian Michael, and Samuel~R Bowman. 2023.
\newblock Gpqa: A graduate-level google-proof q\&a benchmark.
\newblock \emph{arXiv preprint arXiv:2311.12022}.

\bibitem[{Rodriguez et~al.(2021{\natexlab{a}})Rodriguez, Barrow, Hoyle, Lalor, Jia, and Boyd-Graber}]{rodriguez-etal-2021-evaluation}
Pedro Rodriguez, Joe Barrow, Alexander~Miserlis Hoyle, John~P. Lalor, Robin Jia, and Jordan Boyd-Graber. 2021{\natexlab{a}}.
\newblock \href {https://doi.org/10.18653/v1/2021.acl-long.346} {Evaluation examples are not equally informative: How should that change {NLP} leaderboards?}
\newblock In \emph{Proceedings of the 59th Annual Meeting of the Association for Computational Linguistics and the 11th International Joint Conference on Natural Language Processing (Volume 1: Long Papers)}, pages 4486--4503, Online. Association for Computational Linguistics.

\bibitem[{Rodriguez et~al.(2021{\natexlab{b}})Rodriguez, Barrow, Hoyle, Lalor, Jia, and Boyd-Graber}]{rodriguez2021evaluation}
Pedro Rodriguez, Joe Barrow, Alexander~Miserlis Hoyle, John~P Lalor, Robin Jia, and Jordan Boyd-Graber. 2021{\natexlab{b}}.
\newblock Evaluation examples are not equally informative: How should that change nlp leaderboards?
\newblock In \emph{Proceedings of the 59th Annual Meeting of the Association for Computational Linguistics and the 11th International Joint Conference on Natural Language Processing (Volume 1: Long Papers)}, pages 4486--4503.

\bibitem[{Ross et~al.(2024)Ross, Hall, Romero-Soriano, and Williams}]{ross-etal-2024-what}
Candace Ross, Melissa Hall, Adriana Romero-Soriano, and Adina Williams. 2024.
\newblock \href {https://openreview.net/forum?id=LFfktMPAci} {What makes a good metric? evaluating automatic metrics for text-to-image consistency}.
\newblock In \emph{First Conference on Language Modeling}.

\bibitem[{R{\"o}ttger et~al.(2024)R{\"o}ttger, Hofmann, Pyatkin, Hinck, Kirk, Sch{\"u}tze, and Hovy}]{rottger2024political}
Paul R{\"o}ttger, Valentin Hofmann, Valentina Pyatkin, Musashi Hinck, Hannah~Rose Kirk, Hinrich Sch{\"u}tze, and Dirk Hovy. 2024.
\newblock Political compass or spinning arrow? towards more meaningful evaluations for values and opinions in large language models.
\newblock \emph{arXiv preprint arXiv:2402.16786}.

\bibitem[{Saxon et~al.(2024)Saxon, Holtzman, West, Wang, and Saphra}]{saxon-etal-2024-benchmarks}
Michael Saxon, Ari Holtzman, Peter West, William~Yang Wang, and Naomi Saphra. 2024.
\newblock \href {https://openreview.net/forum?id=bttKwCZDkm} {Benchmarks as microscopes: A call for model metrology}.
\newblock In \emph{First Conference on Language Modeling}.

\bibitem[{Shah et~al.(2020)Shah, Gupta, and Roth}]{shah-etal-2020-expect}
Krunal Shah, Nitish Gupta, and Dan Roth. 2020.
\newblock \href {https://doi.org/10.18653/v1/2020.findings-emnlp.317} {What do we expect from multiple-choice {QA} systems?}
\newblock In \emph{Findings of the Association for Computational Linguistics: EMNLP 2020}, pages 3547--3553, Online. Association for Computational Linguistics.

\bibitem[{Singh et~al.(2024)Singh, Kocyigit, Poulton, Esiobu, Lomeli, Szilvasy, and Hupkes}]{singh2024evaluation}
Aaditya~K Singh, Muhammed~Yusuf Kocyigit, Andrew Poulton, David Esiobu, Maria Lomeli, Gergely Szilvasy, and Dieuwke Hupkes. 2024.
\newblock Evaluation data contamination in llms: how do we measure it and (when) does it matter?
\newblock \emph{arXiv preprint arXiv:2411.03923}.

\bibitem[{Sinha et~al.(2021)Sinha, Parthasarathi, Pineau, and Williams}]{sinha-etal-2021-unnatural}
Koustuv Sinha, Prasanna Parthasarathi, Joelle Pineau, and Adina Williams. 2021.
\newblock \href {https://doi.org/10.18653/v1/2021.acl-long.569} {{UnNatural} {L}anguage {I}nference}.
\newblock In \emph{Proceedings of the 59th Annual Meeting of the Association for Computational Linguistics and the 11th International Joint Conference on Natural Language Processing (Volume 1: Long Papers)}, pages 7329--7346, Online. Association for Computational Linguistics.

\bibitem[{Su et~al.(2023)Su, Shi, Kasai, Wang, Hu, Ostendorf, Yih, Smith, Zettlemoyer, and Yu}]{su-etal-2023-one}
Hongjin Su, Weijia Shi, Jungo Kasai, Yizhong Wang, Yushi Hu, Mari Ostendorf, Wen-tau Yih, Noah~A. Smith, Luke Zettlemoyer, and Tao Yu. 2023.
\newblock \href {https://doi.org/10.18653/v1/2023.findings-acl.71} {One embedder, any task: Instruction-finetuned text embeddings}.
\newblock In \emph{Findings of the Association for Computational Linguistics: ACL 2023}, pages 1102--1121, Toronto, Canada. Association for Computational Linguistics.

\bibitem[{Sugawara et~al.(2020)Sugawara, Stenetorp, Inui, and Aizawa}]{sugawara2020assessing}
Saku Sugawara, Pontus Stenetorp, Kentaro Inui, and Akiko Aizawa. 2020.
\newblock Assessing the benchmarking capacity of machine reading comprehension datasets.
\newblock In \emph{Proceedings of the AAAI Conference on Artificial Intelligence}, volume~34, pages 8918--8927.

\bibitem[{Sun et~al.(2021)Sun, Xu, and Suominen}]{sun2021analyzing}
Haozhan Sun, Chenchen Xu, and Hanna Suominen. 2021.
\newblock Analyzing the granularity and cost of annotation in clinical sequence labeling.
\newblock \emph{arXiv preprint arXiv:2108.09913}.

\bibitem[{Swayamdipta et~al.(2020)Swayamdipta, Schwartz, Lourie, Wang, Hajishirzi, Smith, and Choi}]{swayamdipta-etal-2020-dataset}
Swabha Swayamdipta, Roy Schwartz, Nicholas Lourie, Yizhong Wang, Hannaneh Hajishirzi, Noah~A. Smith, and Yejin Choi. 2020.
\newblock \href {https://doi.org/10.18653/v1/2020.emnlp-main.746} {Dataset cartography: Mapping and diagnosing datasets with training dynamics}.
\newblock In \emph{Proceedings of the 2020 Conference on Empirical Methods in Natural Language Processing (EMNLP)}, pages 9275--9293, Online. Association for Computational Linguistics.

\bibitem[{Talat et~al.(2022)Talat, Blix, Valvoda, Ganesh, Cotterell, and Williams}]{talat-etal-2022-machine}
Zeerak Talat, Hagen Blix, Josef Valvoda, Maya~Indira Ganesh, Ryan Cotterell, and Adina Williams. 2022.
\newblock \href {https://doi.org/10.18653/v1/2022.naacl-main.56} {On the machine learning of ethical judgments from natural language}.
\newblock In \emph{Proceedings of the 2022 Conference of the North American Chapter of the Association for Computational Linguistics: Human Language Technologies}, pages 769--779, Seattle, United States. Association for Computational Linguistics.

\bibitem[{Talmor et~al.(2019)Talmor, Herzig, Lourie, and Berant}]{talmor-etal-2019-commonsenseqa}
Alon Talmor, Jonathan Herzig, Nicholas Lourie, and Jonathan Berant. 2019.
\newblock \href {https://doi.org/10.18653/v1/N19-1421} {{C}ommonsense{QA}: A question answering challenge targeting commonsense knowledge}.
\newblock In \emph{Proceedings of the 2019 Conference of the North {A}merican Chapter of the Association for Computational Linguistics: Human Language Technologies, Volume 1 (Long and Short Papers)}, pages 4149--4158, Minneapolis, Minnesota. Association for Computational Linguistics.

\bibitem[{Thompson et~al.(2020)Thompson, Wright, Bissett, and Poldrack}]{thompson-etal-2020-dataset}
William~Hedley Thompson, Jessey Wright, Patrick~G Bissett, and Russell~A Poldrack. 2020.
\newblock Dataset decay and the problem of sequential analyses on open datasets.
\newblock \emph{Elife}, 9:e53498.

\bibitem[{Treviso et~al.(2023)Treviso, Lee, Ji, van Aken, Cao, Ciosici, Hassid, Heafield, Hooker, Raffel, Martins, Martins, Forde, Milder, Simpson, Slonim, Dodge, Strubell, Balasubramanian, Derczynski, Gurevych, and Schwartz}]{treviso-etal-2023-efficient}
Marcos Treviso, Ji-Ung Lee, Tianchu Ji, Betty van Aken, Qingqing Cao, Manuel~R. Ciosici, Michael Hassid, Kenneth Heafield, Sara Hooker, Colin Raffel, Pedro~H. Martins, Andr{\'e} F.~T. Martins, Jessica~Zosa Forde, Peter Milder, Edwin Simpson, Noam Slonim, Jesse Dodge, Emma Strubell, Niranjan Balasubramanian, Leon Derczynski, Iryna Gurevych, and Roy Schwartz. 2023.
\newblock \href {https://doi.org/10.1162/tacl_a_00577} {Efficient methods for natural language processing: A survey}.
\newblock \emph{Transactions of the Association for Computational Linguistics}, 11:826--860.

\bibitem[{Vania et~al.(2021)Vania, Htut, Huang, Mungra, Pang, Phang, Liu, Cho, and Bowman}]{vania-etal-2021-comparing}
Clara Vania, Phu~Mon Htut, William Huang, Dhara Mungra, Richard~Yuanzhe Pang, Jason Phang, Haokun Liu, Kyunghyun Cho, and Samuel~R. Bowman. 2021.
\newblock \href {https://doi.org/10.18653/v1/2021.acl-long.92} {Comparing test sets with item response theory}.
\newblock In \emph{Proceedings of the 59th Annual Meeting of the Association for Computational Linguistics and the 11th International Joint Conference on Natural Language Processing (Volume 1: Long Papers)}, pages 1141--1158, Online. Association for Computational Linguistics.

\bibitem[{Varshney et~al.(2022)Varshney, Mishra, and Baral}]{varshney-etal-2022-ildae}
Neeraj Varshney, Swaroop Mishra, and Chitta Baral. 2022.
\newblock \href {https://doi.org/10.18653/v1/2022.acl-long.240} {{ILDAE}: Instance-level difficulty analysis of evaluation data}.
\newblock In \emph{Proceedings of the 60th Annual Meeting of the Association for Computational Linguistics (Volume 1: Long Papers)}, pages 3412--3425, Dublin, Ireland. Association for Computational Linguistics.

\bibitem[{Vivek et~al.(2024)Vivek, Ethayarajh, Yang, and Kiela}]{vivek-etal-2024-anchor}
Rajan Vivek, Kawin Ethayarajh, Diyi Yang, and Douwe Kiela. 2024.
\newblock \href {https://aclanthology.org/2024.eacl-long.95/} {Anchor points: Benchmarking models with much fewer examples}.
\newblock In \emph{Proceedings of the 18th Conference of the European Chapter of the Association for Computational Linguistics (Volume 1: Long Papers)}, pages 1576--1601, St. Julian{'}s, Malta. Association for Computational Linguistics.

\bibitem[{Wang et~al.(2024{\natexlab{a}})Wang, Zhao, Qiang, Qin, and Liu}]{wang2024beyond}
Haochun Wang, Sendong Zhao, Zewen Qiang, Bing Qin, and Ting Liu. 2024{\natexlab{a}}.
\newblock Beyond the answers: Reviewing the rationality of multiple choice question answering for the evaluation of large language models.
\newblock \emph{arXiv preprint arXiv:2402.01349}.

\bibitem[{Wang et~al.(2024{\natexlab{b}})Wang, Ma, Zhang, Ni, Chandra, Guo, Ren, Arulraj, He, Jiang et~al.}]{wang2024mmlu}
Yubo Wang, Xueguang Ma, Ge~Zhang, Yuansheng Ni, Abhranil Chandra, Shiguang Guo, Weiming Ren, Aaran Arulraj, Xuan He, Ziyan Jiang, et~al. 2024{\natexlab{b}}.
\newblock Mmlu-pro: A more robust and challenging multi-task language understanding benchmark.
\newblock \emph{arXiv preprint arXiv:2406.01574}.

\bibitem[{Wei et~al.(2024)Wei, Wu, Huang, and Chen}]{wei-etal-2024-unveiling}
Sheng-Lun Wei, Cheng-Kuang Wu, Hen-Hsen Huang, and Hsin-Hsi Chen. 2024.
\newblock \href {https://aclanthology.org/2024.findings-acl.333} {Unveiling selection biases: Exploring order and token sensitivity in large language models}.
\newblock In \emph{Findings of the Association for Computational Linguistics ACL 2024}, pages 5598--5621, Bangkok, Thailand and virtual meeting. Association for Computational Linguistics.

\bibitem[{Xu et~al.(2024)Xu, Lou, Du, Mahzoon, Talebianaraki, Zhou, Garrison, Vucetic, and Yin}]{xu2024llms}
Hanzi Xu, Renze Lou, Jiangshu Du, Vahid Mahzoon, Elmira Talebianaraki, Zhuoan Zhou, Elizabeth Garrison, Slobodan Vucetic, and Wenpeng Yin. 2024.
\newblock Llms' classification performance is overclaimed.
\newblock \emph{arXiv preprint arXiv:2406.16203}.

\bibitem[{Yang et~al.(2024)Yang, Yang, Hui, Zheng, Yu, Zhou, Li, Li, Liu, Huang et~al.}]{qwen2}
An~Yang, Baosong Yang, Binyuan Hui, Bo~Zheng, Bowen Yu, Chang Zhou, Chengpeng Li, Chengyuan Li, Dayiheng Liu, Fei Huang, et~al. 2024.
\newblock Qwen2 technical report.
\newblock \emph{arXiv preprint arXiv:2407.10671}.

\bibitem[{Zheng et~al.(2024)Zheng, Zhou, Meng, Zhou, and Huang}]{zheng2024large}
Chujie Zheng, Hao Zhou, Fandong Meng, Jie Zhou, and Minlie Huang. 2024.
\newblock \href {https://openreview.net/forum?id=shr9PXz7T0} {Large language models are not robust multiple choice selectors}.
\newblock In \emph{The Twelfth International Conference on Learning Representations}.

\bibitem[{Zheng et~al.(2023)Zheng, Chiang, Sheng, Zhuang, Wu, Zhuang, Lin, Li, Li, Xing et~al.}]{zheng2023judging}
Lianmin Zheng, Wei-Lin Chiang, Ying Sheng, Siyuan Zhuang, Zhanghao Wu, Yonghao Zhuang, Zi~Lin, Zhuohan Li, Dacheng Li, Eric Xing, et~al. 2023.
\newblock Judging llm-as-a-judge with mt-bench and chatbot arena.
\newblock \emph{Advances in Neural Information Processing Systems}, 36:46595--46623.

\bibitem[{Zong et~al.(2024)Zong, YU, Zhao, Chavhan, and Hospedales}]{zong2024fool}
Yongshuo Zong, Yist~Tingyang YU, Bingchen Zhao, Ruchika Chavhan, and Timothy Hospedales. 2024.
\newblock \href {https://openreview.net/forum?id=H8Qg1IIMaR} {Fool your large (vision and) language models with embarrassingly simple permutations}.

\end{thebibliography}
\bibstyle{acl_natbib}

\clearpage
\appendix

\section{Appendix}
\label{sec:appendix}


\subsection{Finding Threshold for Similar Examples} \label{subsec:threshold_analysis}

To identify similar examples, we analyze cosine distances between SentenceBert embeddings of the examples and identify the first local maxima from the distribution (see Figure \ref{fig:sbert_mmlu}). Notably, we found that considering only the $100$ nearest neighbors for each example is sufficient to determine the threshold accurately (Figure \ref{fig:all_mmlu_dist}). Our analysis, detailed in the Appendix, shows that threshold values obtained using different $k$ values for $k$-nearest neighbors converge after 100 neighbors, aligning closely with results from the entire dataset. For computational efficiency, we therefore use $100$ nearest neighbors to establish the threshold value.

\subsection{Attempt to Identify Wrong Ground Truth}
Benchmark datasets often contain wrong ground truths. In an attempt to identify such examples, we tried an algorithmic framework. We hypothesized that if all top-performing models icorrectly predicted an example with very high confidence (>0.8), then that example would likely be wrongly annotated. However, upon applying this method to the MMLU dataset, our hypothesis didn't work well. Approximately 1.4\% of MMLU examples were filtering using this criteria, but manual inspection revealed that nearly half of these flagged examples actually had correct ground truth annotations, despite unanimous high-confidence wrong predictions by the models. Thus we didn't include this step in \SMART as it would require human involvement and may not provide sufficient efficiency gains to justify the additional resource expenditure.

\begin{figure*}[t]
  \centering
  \subfloat[k=100]{\includegraphics[width=0.5\textwidth]{figures/sbert_mmlu_knn=100.jpg}} \hfill
  \subfloat[k=500]{\includegraphics[width=0.5\textwidth]{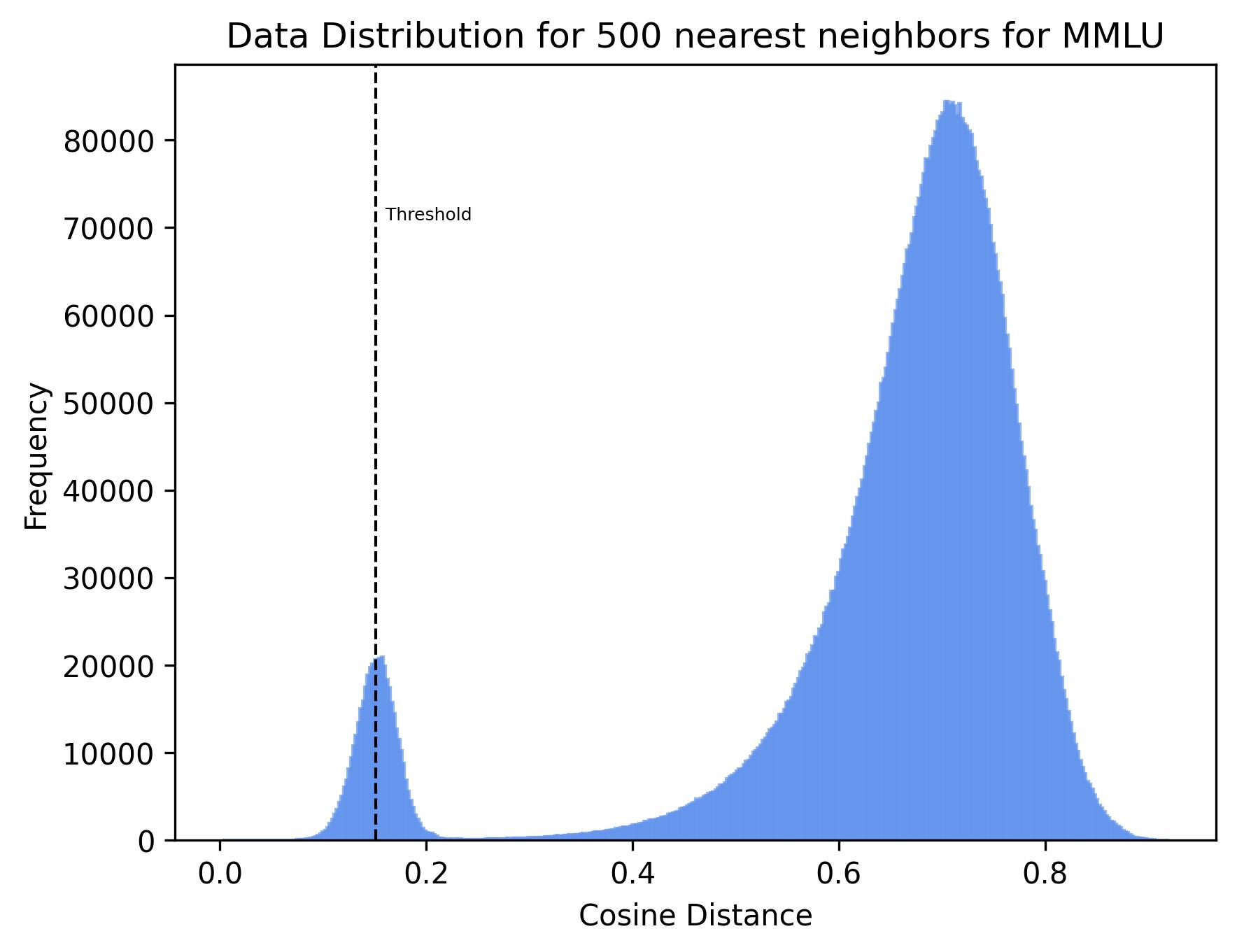}} \hfill \\
  \subfloat[k=1000]{\includegraphics[width=0.5\textwidth]{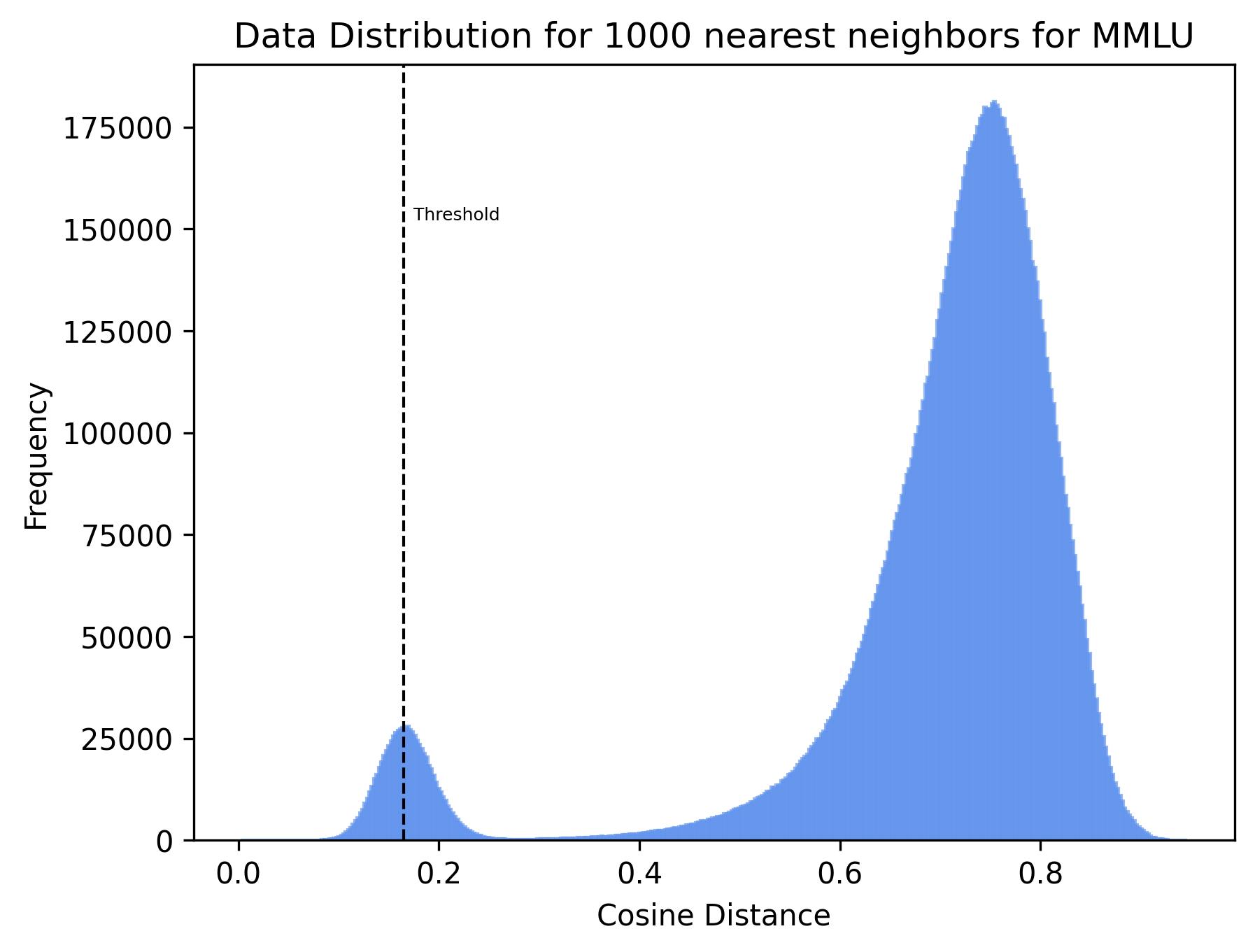}} \hfill
  \subfloat[k=5000]{\includegraphics[width=0.5\textwidth, height=6.1cm]{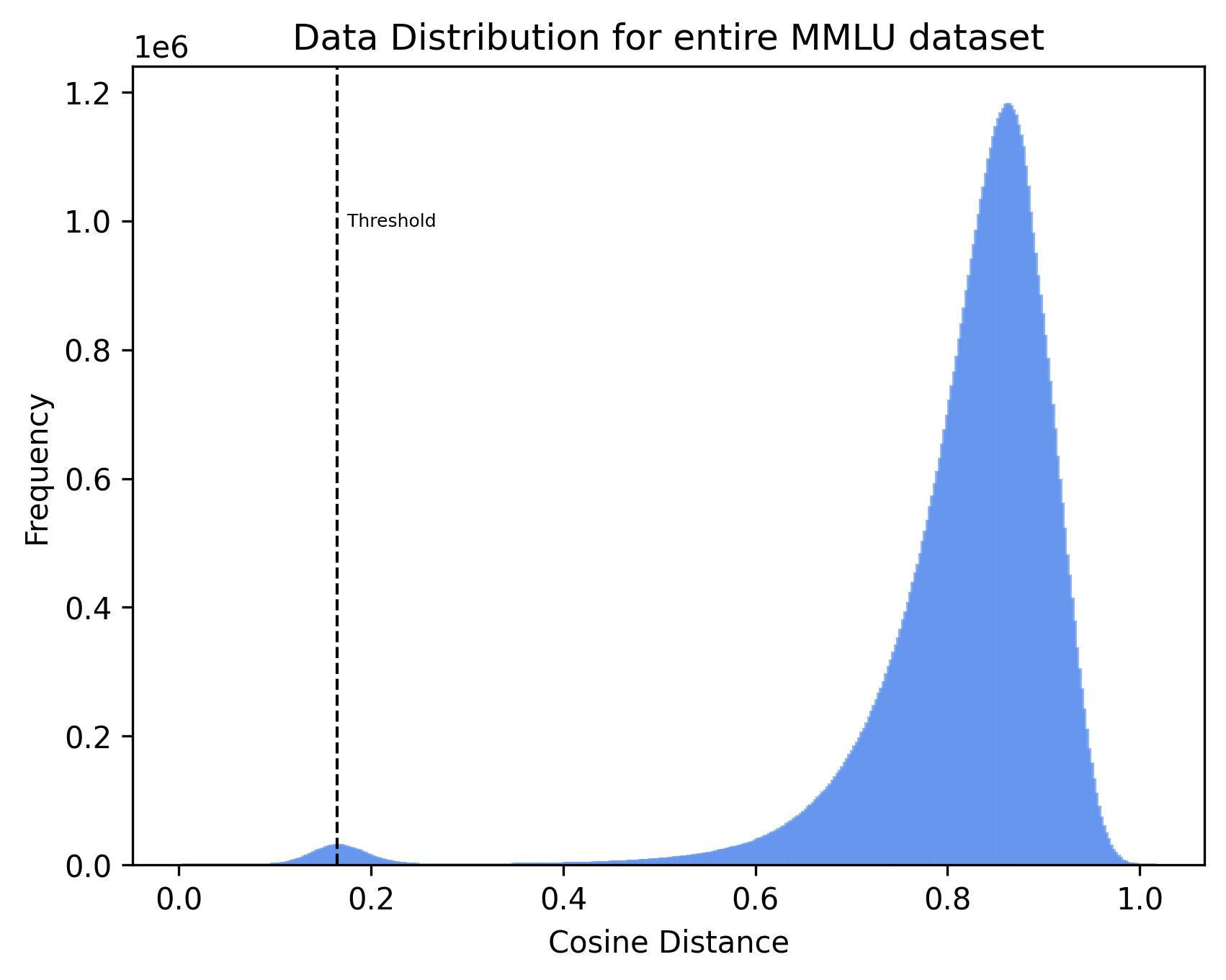}} \hfill
  \caption{Distribution of cosine distances between SentenceBert embeddings for different values of \textit{k} nearest neighbors for each example in the MMLU dataset. We find that threshold values obtained using different $k$ values for $k$-nearest neighbors converge after 100 neighbors.}
  \label{fig:all_mmlu_dist}
\end{figure*}

\subsection{Exact Match Duplicates} \label{subsec:exact_match}
Dataset creators often preprocess their datasets to remove duplicate examples before releasing them. Nevertheless, exact-match duplicates can still be present. Our analysis reveals that 1.2\% of the MMLU dataset consists of identical questions, while ARC contains 0.2\% duplicates. In contrast, CommonsenseQA does not have any exact duplicates, with a rate of 0\%.

\subsection{Category Wise Results}\label{appsubsec:category}

MMLU has 57 different categories and we found that different categories in MMLU are affected differently by \SMART. The percentage of examples removed by \SMART for each category is given in Table \ref{tab:mmlu_categories}.

\subsection{LLM Based Embedding for Similar Examples}
In addition to using LLM2Vec embeddings, we also used the last token's LLM embeddings as representation of entire sentences \cite{neelakantan2022text, ma2024fine}
The results as shown in Table \ref{tab:embedding_comparison}, shows that SentenceBert embeddings are very similar to LLM based embeddings for identifying and removing similar examples.

\begin{table}[h!]
\centering
\begin{tabular}{l|c}
\toprule
\textbf{Embedding Pair} & \textbf{Percentage Overlap}\\
\midrule
SBert and Llama-3-70B & 94.7 \\
SBert and Qwen2-72B & 95.2\\
SBert and Llama-3-8B & 95.8\\
\bottomrule
\end{tabular}
\caption{
Percentage of examples with extremely high semantic similarity, based on a distance in the embedding space smaller than a $\delta$. We define the process for computing $\delta$ in Section \ref{sec:threshold}.
}

\label{tab:embedding_comparison}
\end{table}

\subsection{Computation Resources}
For all experiments for this work, we utilized A100 80GB GPUs. Depending on the model evaluated, we used 1,2,3 or 4 GPUs for inferences. These GPUs were assembled in a cluster of 8 GPUs in a node. The cumulative computing time required to evaluate all the language models and complete the experiments amounted to approximately 2000 GPU hours. We also used LLMs for coding assistance for building our codebase.

\subsection{Detailed Model Results}

Table \ref{tab:arc_accuracy}, \ref{tab:mmlu_accuracy}, and \ref{tab:cqa_comparison} shows accuracy for all tested models on ARC, MMLU, and CommonsenseQA dataset and their SMART filtered versions.

\begin{table*}[ht]
\centering
\small
\begin{tabular}{|l|r|r|r|}
   \toprule
    Category Name & Original \# of examples & SMART filtered examples & \makecell{Percentage of \\ examples removed} \\
    \midrule
    Moral Scenarios & 895 & 0 & 100.00\% \\
    High School Government And Politics & 193 & 52 & 73.06\% \\
    Us Foreign Policy & 100 & 31 & 69.00\% \\
    High School Psychology & 545 & 182 & 66.61\% \\
    Miscellaneous & 783 & 274 & 65.01\% \\
    Marketing & 234 & 86 & 63.25\% \\
    Sociology & 201 & 76 & 62.19\% \\
    International Law & 121 & 46 & 61.98\% \\
    High School World History & 237 & 91 & 61.60\% \\
    World Religions & 171 & 70 & 59.06\% \\
    High School Us History & 204 & 85 & 58.33\% \\
    High School Geography & 198 & 83 & 58.08\% \\
    Medical Genetics & 100 & 43 & 57.00\% \\
    High School Biology & 310 & 134 & 56.77\% \\
    High School European History & 165 & 72 & 56.36\% \\
    Logical Fallacies & 163 & 72 & 55.83\% \\
    Management & 103 & 46 & 55.34\% \\
    College Medicine & 173 & 80 & 53.76\% \\
    Prehistory & 324 & 155 & 52.16\% \\
    College Biology & 144 & 70 & 51.39\% \\
    High School Microeconomics & 238 & 121 & 49.16\% \\
    Clinical Knowledge & 265 & 136 & 48.68\% \\
    Jurisprudence & 108 & 56 & 48.15\% \\
    Astronomy & 152 & 79 & 48.03\% \\
    Security Studies & 245 & 129 & 47.35\% \\
    Computer Security & 100 & 54 & 46.00\% \\
    Human Sexuality & 131 & 72 & 45.04\% \\
    High School Computer Science & 100 & 55 & 45.00\% \\
    Professional Psychology & 612 & 337 & 44.93\% \\
    Nutrition & 306 & 171 & 44.12\% \\
    High School Macroeconomics & 390 & 228 & 41.54\% \\
    Public Relations & 110 & 67 & 39.09\% \\
    Human Aging & 223 & 138 & 38.12\% \\
    Business Ethics & 100 & 62 & 38.00\% \\
    Professional Medicine & 272 & 173 & 36.40\% \\
    Philosophy & 311 & 199 & 36.01\% \\
    Moral Disputes & 346 & 223 & 35.55\% \\
    Anatomy & 135 & 89 & 34.07\% \\
    Conceptual Physics & 235 & 163 & 30.64\% \\
    College Physics & 102 & 75 & 26.47\% \\
    Electrical Engineering & 145 & 107 & 26.21\% \\
    Virology & 166 & 127 & 23.49\% \\
    High School Chemistry & 203 & 158 & 22.17\% \\
    College Chemistry & 100 & 79 & 21.00\% \\
    High School Statistics & 216 & 173 & 19.91\% \\
    Machine Learning & 112 & 90 & 19.64\% \\
    Econometrics & 114 & 93 & 18.42\% \\
    College Computer Science & 100 & 83 & 17.00\% \\
    Professional Accounting & 282 & 244 & 13.48\% \\
    Elementary Mathematics & 378 & 330 & 12.70\% \\
    Formal Logic & 126 & 110 & 12.70\% \\
    Professional Law & 1534 & 1340 & 12.65\% \\
    High School Physics & 151 & 143 & 5.30\% \\
    High School Mathematics & 270 & 259 & 4.07\% \\
    Abstract Algebra & 100 & 96 & 4.00\% \\
    Global Facts & 100 & 96 & 4.00\% \\
    College Mathematics & 100 & 97 & 3.00\% \\
\midrule
\end{tabular}
\caption{Results of each of the 57 categories of \SMART on MMLU dataset. Different categories are affected differently.}
\label{tab:mmlu_categories}
\end{table*}

\begin{table*}[ht]
\centering
\begin{tabular}{|l|r|r|}
\midrule
\textbf{Model Name} & \textbf{Accuracy on ARC} & \textbf{Accuracy on ARC-SMART} \\
\midrule
Qwen2-72B-Instruct & 0.939 & 0.83 \\
Meta-Llama-3.1-70B-Instruct & 0.934 & 0.819 \\
Meta-Llama-3-70B-Instruct & 0.933 & 0.819 \\
gemma-2-27b-it & 0.925 & 0.788 \\
Phi-3-medium-4k-instruct & 0.923 & 0.781 \\
Phi-3.5-MoE-instruct & 0.92 & 0.785 \\
Mixtral-8x22B-Instruct-v0.1 & 0.916 & 0.762 \\
gemma-2-9b-it & 0.91 & 0.757 \\
Yi-34B-Chat & 0.909 & 0.745 \\
Qwen1.5-32B-Chat & 0.907 & 0.752 \\
dbrx-instruct & 0.905 & 0.732 \\
Yi-34B & 0.902 & 0.724 \\
Yi-1.5-9B-Chat & 0.894 & 0.728 \\
Meta-Llama-3-8B-Instruct & 0.883 & 0.721 \\
Qwen2-7B-Instruct & 0.882 & 0.697 \\
Mixtral-8x7B-Instruct-v0.1 & 0.876 & 0.681 \\
Mixtral-8x7B-v0.1 & 0.876 & 0.688 \\
internlm2\_5-20b-chat & 0.875 & 0.675 \\
internlm2\_5-7b-chat & 0.861 & 0.647 \\
Llama-2-70b-hf & 0.855 & 0.644 \\
gemma-7b & 0.837 & 0.611 \\
Mistral-7B-v0.3 & 0.802 & 0.563 \\
Mistral-7B-Instruct-v0.2 & 0.793 & 0.581 \\
Qwen-7B-Chat & 0.764 & 0.518 \\
gemma-7b-it & 0.738 & 0.531 \\
Qwen-7B & 0.733 & 0.476 \\
falcon-40b & 0.722 & 0.51 \\
falcon-40b-instruct & 0.722 & 0.507 \\
OLMo-1.7-7B-hf & 0.639 & 0.435 \\
\midrule
\end{tabular}
\caption{Accuracy of different models on ARC and ARC-\SMART datasets. Almost all the models have similar ranking among the two versions of ARC.}
\label{tab:arc_accuracy}
\end{table*}

\begin{table*}[ht]
\centering
\begin{tabular}{|l|r|r|}
\midrule
\textbf{Model Name} & \textbf{Accuracy on MMLU} & \textbf{Accuracy on MMLU-SMART} \\
\midrule
Qwen2-72B-Instruct & 0.841 & 0.743 \\
Meta-Llama-3.1-70B-Instruct & 0.823 & 0.714 \\
Meta-Llama-3-70B-Instruct & 0.803 & 0.692 \\
Phi-3.5-MoE-instruct & 0.787 & 0.67 \\
Phi-3-medium-4k-instruct & 0.781 & 0.656 \\
Mixtral-8x22B-Instruct-v0.1 & 0.779 & 0.653 \\
Yi-1.5-34B-Chat & 0.764 & 0.634 \\
gemma-2-27b-it & 0.762 & 0.639 \\
Yi-34B & 0.758 & 0.624 \\
Qwen1.5-32B-Chat & 0.754 & 0.615 \\
Yi-34B-Chat & 0.747 & 0.603 \\
dbrx-instruct & 0.732 & 0.6 \\
gemma-2-9b-it & 0.725 & 0.588 \\
internlm2\_5-7b-chat & 0.711 & 0.568 \\
Mixtral-8x7B-Instruct-v0.1 & 0.706 & 0.565 \\
Mixtral-8x7B-v0.1 & 0.704 & 0.568 \\
Qwen2-7B-Instruct & 0.702 & 0.564 \\
internlm2\_5-20b-chat & 0.701 & 0.567 \\
Yi-1.5-9B-Chat & 0.699 & 0.556 \\
Llama-2-70b-hf & 0.69 & 0.544 \\
Meta-Llama-3-8B-Instruct & 0.664 & 0.505 \\
gemma-7b & 0.644 & 0.492 \\
Mistral-7B-v0.3 & 0.626 & 0.468 \\
Mistral-7B-Instruct-v0.2 & 0.593 & 0.441 \\
Qwen-7B & 0.574 & 0.426 \\
Qwen-7B-Chat & 0.563 & 0.415 \\
falcon-40b & 0.558 & 0.412 \\
falcon-40b-instruct & 0.547 & 0.402 \\
OLMo-1.7-7B-hf & 0.521 & 0.381 \\
gemma-7b-it & 0.517 & 0.389 \\
\midrule
\end{tabular}
\caption{Model Accuracy Comparison on MMLU and MMLU-\SMART. Almost all the models have similar ranking among the two versions of MMLU.}
\label{tab:mmlu_accuracy}
\end{table*}

\begin{table*}[ht]
\centering
\begin{tabular}{|l|r|r|}
\midrule
Model Name & Accuracy on CommonsenseQA & \makecell{Accuracy on \\ CommonsenseQA-SMART} \\
\midrule
Qwen2-72B-Instruct & 0.885 & 0.845 \\
Yi-1.5-34B-Chat & 0.833 & 0.776 \\
Meta-Llama-3-70B-Instruct & 0.833 & 0.771 \\
Qwen1.5-32B-Chat & 0.830 & 0.767 \\
Meta-Llama-3.1-70B-Instruct & 0.815 & 0.741 \\
Phi-3.5-MoE-instruct & 0.809 & 0.739 \\
gemma-2-9b-it & 0.808 & 0.733 \\
Qwen2-7B-Instruct & 0.804 & 0.724 \\
Phi-3-medium-4k-instruct & 0.797 & 0.722 \\
Yi-34B & 0.797 & 0.718 \\
gemma-2-27b-it & 0.794 & 0.719 \\
Yi-34B-Chat & 0.791 & 0.712 \\
Yi-1.5-9B-Chat & 0.790 & 0.718 \\
internlm2\_5-7b-chat & 0.789 & 0.714 \\
dbrx-instruct & 0.786 & 0.704 \\
internlm2\_5-20b-chat & 0.777 & 0.695 \\
Meta-Llama-3-8B-Instruct & 0.765 & 0.680 \\
Mixtral-8x22B-Instruct-v0.1 & 0.759 & 0.672 \\
OLMo-1.7-7B-hf & 0.739 & 0.670 \\
Mixtral-8x7B-Instruct-v0.1 & 0.706 & 0.600 \\
gemma-7b-it & 0.688 & 0.594 \\
Qwen-7B & 0.683 & 0.586 \\
Mistral-7B-Instruct-v0.2 & 0.682 & 0.590 \\
falcon-40b-instruct & 0.678 & 0.579 \\
Qwen-7B-Chat & 0.652 & 0.557 \\
gemma-7b & 0.646 & 0.551 \\
Mistral-7B-v0.3 & 0.603 & 0.499 \\
Llama-2-70b-hf & 0.583 & 0.465 \\
Mixtral-8x7B-v0.1 & 0.581 & 0.468 \\
falcon-40b & 0.555 & 0.446 \\
    \midrule
\end{tabular}
\caption{Accuracy comparison of models on CommonsenseQA and CommonsenseQA-\SMART. Almost all the models have similar ranking among the two versions of CommonsenseQA.}
\label{tab:cqa_comparison}
\end{table*}

\end{document}